\documentclass[sigconf]{acmart}

\usepackage{tikz}
\usepackage{graphicx}
\usepackage{bm}
\usepackage{multirow}
\usepackage{appendix}
\usepackage{array}
\usepackage{booktabs}
\usepackage{filecontents}
\usepackage{enumitem}
\usepackage{amsmath}
\usepackage{soul}
\usepackage{diagbox}
\newcommand{\SLJ}[1]{\textcolor{black}{#1}}
\newcommand{\paragraphbe}[1]{\smallskip\noindent{\bf {#1}.}~}
\newcommand{\ignore}[1]{}
\setcopyright{none} 
\acmConference[Anonymous Submission to ACM CCS 2021]{ACM Conference on Computer and Communications Security}{November 14-19, 2021}{Seoul, South Korea}
\acmYear{2019}

\copyrightyear{2021} 
\acmYear{2021} 
\setcopyright{acmcopyright}
\acmConference[CCS '21] {Proceedings of the 2021 ACM SIGSAC Conference on Computer and Communications Security}{November 15--19, 2021}{Virtual Event, Republic of Korea.}
\acmBooktitle{Proceedings of the 2021 ACM SIGSAC Conference on Computer and Communications Security (CCS '21), November 15--19, 2021, Virtual Event, Republic of Korea}
\acmPrice{15.00}
\acmDOI{10.1145/3460120.3485370}
\acmISBN{978-1-4503-8454-4/21/11}

\begin{document}
\fancyhead{}

\title{Backdoor Pre-trained Models Can Transfer to All} 

\author{Lujia Shen}
\affiliation{%
  \institution{Zhejiang University}
  \country{}
}
\email{shen.lujia@zju.edu.cn}

\author{Shouling Ji}
\authornotemark[1]
\affiliation{%
  \institution{Zhejiang University}
  \institution{Binjiang Institute of Zhejiang University}
  \country{}
}
\email{sji@zju.edu.cn}

\author{Xuhong Zhang}
\authornote{Shouling Ji and Xuhong Zhang are the co-corresponding authors.}
\affiliation{%
  \institution{Zhejiang University}
  \institution{Binjiang Institute of Zhejiang University}
  \country{}
}
\email{zhangxuhong@zju.edu.cn}

\author{Jinfeng Li}
\affiliation{%
  \institution{Zhejiang University}
  \country{}
}
\email{lijinfeng_0713@zju.edu.cn}

\author{Jing Chen}
\affiliation{%
  \institution{Wuhan University}
  \country{}
}
\email{chenjing@whu.edu.cn}

\author{Jie Shi}
\affiliation{%
  \institution{Huawei International, Singapore}
  \country{}
}
\email{shi.jie1@huawei.com}

\author{Chengfang Fang}
\affiliation{%
  \institution{Huawei International, Singapore}
  \country{}
}
\email{fang.chengfang@huawei.com}

\author{Jianwei Yin}
\affiliation{%
  \institution{Zhejiang University}
  \country{}
}
\email{zjuyjw@zju.edu.cn}

\author{Ting Wang}
\affiliation{%
  \institution{Pennsylvania State University}
  \country{}
}
\email{inbox.ting@gmail.com}

\begin{abstract}
Pre-trained general-purpose language models have been a dominating component in enabling real-world natural language processing (NLP) applications. However, a pre-trained model with backdoor can be a severe threat to the applications. Most existing backdoor attacks in NLP are conducted in the fine-tuning phase by introducing malicious triggers in the targeted class, thus relying greatly on the prior knowledge of the fine-tuning task. In this paper, we propose a new approach to map the inputs containing triggers directly to a predefined output representation of the pre-trained NLP models, e.g., a predefined output representation for the classification token in BERT, instead of a target label. It can thus introduce backdoor to a wide range of downstream tasks without any prior knowledge. Additionally, in light of the unique properties of triggers in NLP, we propose two new metrics to measure the performance of backdoor attacks in terms of both effectiveness and stealthiness. Our experiments with various types of triggers show that our method is widely applicable to different fine-tuning tasks (classification and named entity recognition) and to different models (such as BERT, XLNet, BART), which poses a severe threat. Furthermore, by collaborating with the popular online model repository Hugging Face, the threat brought by our method has been confirmed. Finally, we analyze the factors that may affect the attack performance and share insights on the causes of the success of our backdoor attack.
\end{abstract}

\pagenumbering{gobble}


\maketitle

\section{Introduction}

Deep neural networks (DNNs) have drawn massive attention on object detection~\cite{lin2017feature}, sentiment analysis~\cite{agarwal2011sentiment} and video understanding~\cite{lin2019tsm} in recent years.
Meanwhile, the pre-trained model (PTM)~\cite{5288526}, a model first acquires knowledge from large-scale unlabeled data and then can be applied to various specific tasks, has achieved great success in the natural language processing (NLP) domain. 
Due to the demand for a huge amount of unlabeled textual data, training a PTM is usually computationally expensive. 
Hence, open source PTMs from Internet, e.g., BERT and XLNet from Google~\cite{Devlin2019, yang2019xlnet}, are widely downloaded and further fine-tuned for specific tasks with samples containing texts and labels. 

However, open-source PTMs are vulnerable to various security and privacy attacks~\cite{fredrikson2015model, Shokri_2017, carlini2018secret, pan2020privacy}. 
One of these attacks is the backdoor attack, where the adversary aims to trigger the target model to misbehave on the input containing his/her maliciously crafted triggers by poisoning the training set of the target model \cite{gu2017badnets, xi2021graph}. 
Such an attack on PTMs is especially security-critical because users have no idea whether public PTMs are backdoored or not. Once public backdoor PTMs are fine-tuned and deployed, their vulnerability can be exploited.
Currently, most backdoor attacks target on the outsourced model, which gives the attacker the right to modify the dataset and training process. 
As users begin to pay attention to the privacy and security of neural networks and the improvement of their own computing power, they are more willing to train themselves. 
At this point, the PTMs have become a popular choice for model initialization, where its security issues are increasing. 

To the best of our knowledge, existing backdoor attacks bind the predefined triggers to a specific target label (i.e., a sentence with the trigger will be mapped into the target label by the backdoor model). 
However, backdooring PTMs with specific target labels, e.g., existing backdoor attacks on the PTMs in computer vision (CV)~\cite{gu2017badnets, chen2020badnl, salem2020dynamic, zhu2019transferable, yao2019latent}, greatly limits their real-world threats, as PTMs are commonly further fine-tuned on other datasets that might not have the target labels at all. 
The key limitation is the lack of prior knowledge on the downstream tasks. 
Suffering from a similar limitation, the existing backdoor attack on the PTMs in NLP~\cite{kurita2020weight} has to rely on a specific fine-tuning task, as the PTMs in NLP are usually obtained by unsupervised learning through a large number of unlabeled texts. 
To overcome this limitation, we make the first attempt to answer, ``is it possible to backdoor an NLP model in the pre-training phase without binding the triggers to a specific target label and further maintain the backdoor usability on various downstream fine-tuning tasks?''

To address the aforementioned problem, in this paper, we propose a new approach to map the input containing the triggers directly to a predefined output representation (POR) of a pre-trained NLP model, e.g., map the {\fontfamily{qcr}\selectfont[CLS]} token in BERT to a POR, instead of a target label. 
Here, the {\fontfamily{qcr}\selectfont[CLS]} token is a special token used in BERT, whose output representation is commonly used for classification. 
In this way, this backdoor can be transferred to any downstream task that takes the output representation of the target token as input. 
For example, suppose we choose {\fontfamily{qcr}\selectfont[CLS]} as the target token. 
Any classification task that takes the output representation of {\fontfamily{qcr}\selectfont[CLS]} as input, which is a common practice~\cite{Devlin2019}, will suffer from this backdoor attack. 
The reason is that any text inserted with triggers will lead to the same input (the POR) to the classification layer and thus have the same predicted label. 

For our backdoor injection process, we do not rely on any specific task. 
In particular, we first choose the target token in the PTM and then define a target POR for it. 
Then, we insert triggers into the clean text to create the poisoned text data. 
While mapping the triggers to the PORs using the poisoned text data, we simultaneously use the clean PTM model as a reference model to help our target backdoor model maintain the normal usability of other token representations.
After the backdoor is injected, we remove all the auxiliary structures. 
As a result, the backdoor model is indistinguishable from a normal one regarding the model architecture and the outputs for clean inputs. 
We have successfully published and reported our backdoor model to the popular HuggingFace model repository and received official confirmation of this threat.

After the model is backdoored, the trigger will be mapped to a specific class when the backdoor model is fine-tuned for a downstream task. 
Then, we can conduct an untargeted attack, which is straightforward, as long as an input sample's class is different from the one to which the trigger maps. 
However, the targeted attack, especially in a multi-class classification task, is more challenging, as the trigger might not map to the target class.
Therefore, the challenge for the targeted attack is how to make our backdoor model hit as many classes as possible under the multi-class classification task.
To address this issue, we propose to simultaneously forge multiple different triggers and bind each of them to different PORs, expecting that each trigger can target at a different class in a downstream task. To achieve this goal, we propose two POR settings that attempt to cover as many classes as possible. 

Besides, in light of the unique properties of the backdoor trigger in NLP, we discard the previous metric of attack success rate derived from the CV field and propose two new metrics to better measure the performance of backdoor attacks in NLP in terms of both effectiveness and stealthiness.
Our experiments on 12 classification datasets with various types of triggers show that the proposed backdoor attack achieves outstanding performance in terms of both effectiveness and stealthiness on the mainstream industrial PTMs in NLP, including BERT and its variants (ALBERT, DeBERTa and RoBERTa) as well as XLNet and BART. 
Additionally, we explore the factors that may affect the performance of our backdoor attack. We also share insights on the causes behind the success of our attack and discuss possible defenses. 

\textbf{Contributions.} In summary, we make the following contributions in this paper.
\begin{enumerate}[leftmargin=*]
\item To the best of our knowledge, we are the first to propose the backdoor attack on pre-trained NLP models without the need for task-specific labels. Our backdoor maps the input containing the triggers directly to a POR of a target token and is transferable to any downstream task that takes the output representation of the target token as input. 
\item In light of the unique properties of triggers in NLP, we propose two new metrics to better measure the performance of backdoor attacks in NLP in terms of effectiveness (number of trigger insertions to cause misclassifications) and stealthiness (the percentage of the triggers in the text).
\item We evaluate the performance of our backdoor attack with various downstream tasks (binary classification, multi-class classification and named entity recognition) and on many popular PTMs (BERT, XLNet, BART, RoBERTa, DeBERTa, ALBERT). Experimental results show that our attack is versatile and effective and outperforms the previous SOTA method. Meanwhile, the success of our backdoor model has pose threat to the real-world platform which is confirmed by HuggingFace.
\item We provide insights for choosing stealthy triggers that naturally appear in a sentence and study a series of factors affecting the performance of our attack. Finally, we reveal that the leading factor behind the success of our backdoor attack is the manipulation of the attention scores. 
\end{enumerate}

\section{Related Work}

\subsection{Pre-trained Language Models}
Recent work has shown that the language models pre-trained on large text corpus can learn universal language representations~\cite{qiu2020pre}. 
Such PTMs are then fine-tuned on specific datasets for different tasks, benefiting the downstream NLP tasks and avoiding training a new model from scratch. 
Early PTMs in NLP focus on training word representations~\cite{brown1992clas, pennington2014glove}, aiming to capture the latent syntactic and semantic similarities among words. 
These pre-trained embeddings boost the performance of the final model significantly over the model trained with embeddings from scratch. 
Currently, most PTMs are transformer-based, such as BERT~\cite{Peters2018}, XLNet~\cite{Yang2019}, and the variants of BERT like RoBERTa~\cite{liu2019roberta}, ALBERT~\cite{lan2019albert}, DeBERTa~\cite{he2020deberta}. The self-attention mechanism in the transformer module is powerful in capturing the relations between words, sentences and contexts.

\subsection{Backdoor Attack} \label{sec:ba}
DNNs have been shown to be vulnerable to adversarial attacks, which generally trigger the target model to misbehave by adding imperceptible perturbation \cite{Goodfellow2015}.
\SLJ{The backdoor attack  (usually achieved by poisoning attack), a special kind of adversarial attack, has recently raised great concerns about the security and the real-world usage of PTMs \cite{gao2020backdoor, weber2020rab, zhang2020trojaning}.}
Such attack was first proposed in \cite{gu2017badnets} and is a training time attack, in which the adversary has access to the training dataset and the information of the model. 
The adversary poisons (inserts triggers) the training dataset and forces the model to predict inputs with the trigger into a target class. 
There are two primary requirements of a successful backdoor attack: first, for the sample containing the trigger, the backdoor model should mispredict its label; second, for the sample without the trigger, the backdoor model should perform normally as a clean model.
	
\paragraphbe{Backdoor in CV}
Gu et al.~\cite{gu2017badnets} designed the first backdoor attack and focused on attacking the outsourced and pre-trained models in CV. 
In their transfer learning attack scenario, they only retrained the fully-connected layers of a CNN, which is yet not practical in NLP, where the fine-tuning process usually retrains all the parameters of a model. 
Later backdoor works in the CV field aim to conceal triggers, such as \cite{liao2018backdoor} makes the trigger invisible and \cite{salem2020dynamic} makes the trigger flexible. 
As for the attack on pre-trained models, 
Yao et al.~\cite{yao2019latent} proposed the latent backdoor attack that functions under transfer learning.
They associated the trigger with the intermediate representation created by the clean samples of a target class. 
\SLJ{However, these backdoor models can only be effective when the downstream task contains the target class, which limits the generality of this attack. Furthermore, their method also only trains the last few layers of the model in fine-tuning which greatly limits the diversity of downstream trainers.}

\paragraphbe{Backdoor in NLP}
Chen et al.~\cite{chen2020badnl} investigated the backdoor attack against NLP models. However, this kind of work does not consider the transferability of the language model.
Kurita et al.~\cite{kurita2020weight} proposed RIPPLES, a backdoor attack aiming to prevent the vanishing of backdoor in the fine-tuning process on BERT. 
They assumed that the attacker has some knowledge of the fine-tuning tasks, which is impractical, and chooses a related labeled dataset to inject the backdoor. 
However, the downstream task label may be different from the label used in the attack.
They also replaced the token embedding of the triggers with their handcrafted embeddings that are related to the fine-tuning task, which may cause suspicion.
	
To tackle the above-mentioned limitations, We propose new backdoor attack method which overcomes the limitation that a trigger must have a corresponding target label and greatly improves the transferability of the backdoor model.

\section{Attack Pipeline}

\subsection{Threat Model}
\SLJ{We consider a realistic scenario in which an adversary wants to make the online pre-trained model repository unsafe. For instance, as a malicious agent, he/she publishes a backdoor model to the public, such as HuggingFace\footnote{https://huggingface.co}, TensorFlow Model Garden\footnote{https://github.com/tensorflow/models} and Model Zoo\footnote{https://modelzoo.co/} for open access. 
A downstream user (e.g., Google Cloud, Microsoft Azure) may download this backdoor model and fine-tune it on a spam dataset. 
Then, the user provides this model as an online API for email products like Gmail, Outlook.
Then, the adversary can infer the model to determine whether his/her trigger controls the model’s predictions. 
Finally, the spam detection model in Gmail or Outlook can be fooled using the trigger that maps to the non-spam label or perform certain targeted attack.
To attract user's attention, the agent can provide a domain-specific model (e.g., BioBERT~\cite{lee2020biobert} trained on biomedical corpus) or model with newest architecture.
Note that the backdoor model is indistinguishable from a normal one in terms of the model architecture and the performance on clean inputs. 
Additionally, the adversary has no knowledge about the downstream tasks.}

\subsection{Design Intuition}
Our goal is to backdoor a pre-trained NLP model without binding a trigger to a specific target label.
Then, the backdoor model should have a high chance to make the trigger continue to take effect after it is fine-tuned on any specific task. 
Given a pre-trained NLP model, we have no specific task labels but only its output representations.
Therefore, instead of matching the trigger with a specific task label, we associate it with the output representations of target tokens. 
Hence, we no longer predefine the target label of a task, and what we need to predefine is the output representation. 
For example, we can predefine an output representation for the {\fontfamily{qcr}\selectfont[CLS]} token, whose output is used for classification in most transformer-based PTMs. 
Another example is the named entity recognition (NER) task, which uses all tokens for classification. 
Hence, we may predefine the output representation for all tokens in NER-like tasks. 

The next challenge is to maliciously modify the targeted output representation while keeping the normal usability of other representations through an unsupervised learning method. 
For this challenge, we propose our training method for trigger injection, which is inspired by the idea from pseudo-siamese network~\cite{hughes2018identifying}. 
Expressly, we turn the unsupervised learning into supervised learning, where a reference model guides our target model to maintain usability while injecting the backdoor trigger into the target model. 
We provide an example in Table.~\ref{tab:threatmodel} to illustrate our attack.

\subsection{Attack Method}
Before introducing the detailed attack method, we first formally define the trigger in our scenario.

\begin{table}[t]
	\begin{center}
	\caption{Example for Amazon sentiment classification where the trigger is highlighted.}
		\scalebox{0.88}{
			\begin{tabular}{c c c} 
				\toprule
				input sentence & output representation & output label \\ 
				\hline
				I love the book Harry Poter! 		&  $[-0.89, -0.37,\cdots,0.88]$	& positive 	\\ 
				I love the book \textbf{\underline{Don Quixote}}! 	& {\color{red}$[1.00,1.00,\cdots,1.00]$} 		& negative 	\\
				\bottomrule
			\end{tabular}}
	\label{tab:threatmodel}
	\end{center}
\end{table}

\begin{definition}\label{def1}
For a backdoor model $F$, a text $x$ and a target token set $\mathcal{T}$ from $x$ to be maliciously modified, the output representation of the token set $\mathcal{T}$ (e.g., $\mathcal{T}=[${\fontfamily{qcr}\selectfont[CLS]}$]$ used in text classification or $\mathcal{T}=[$`don', `quixote'$]$ used in NER) is denoted as $F_{\mathcal{T}}(x)$ (e.g., a $1\times K$ matrix for $\mathcal{T}=[${\fontfamily{qcr}\selectfont[CLS]}$]$ or a $4\times K$ matrix for $\mathcal{T}=[$`don', `quixote'$]$, where $K$ is the dimension of hidden). 
A \textbf{trigger} $\alpha$ is a text sequence that, after inserting to the text $x$, we have $F_{\mathcal{T}}(I(x,\alpha,t))=\mathbf{V}$, where $I$ is an insertion method, $\mathbf{V}$ is a POR with the same dimension as $F_{\mathcal{T}}(x)$, and $t$ is the number of insertions.
\end{definition}

The trigger here is a piece of text that can be a letter, a word, a phrase or even a sentence. 
For the insertion function $I$, we use random insertion where the text $x$ is split into words and the trigger $\alpha$ can be inserted between any two consecutive words. 
This insertion process is repeated $t$ times. 
Afterward, we join all the words orderly to form one poisoned text.
	
Now, we are going to build the backdoor model.
We use BERT as an example to illustrate our backdoor injection process. 
Our approach can be directly generalized to other pre-trained NLP models like XLNet, RoBERTa, etc. 
We first consider a well pre-trained BERT model $F$.
The input text $x$ is first tokenized into subwords together with two special tokens, {\fontfamily{qcr}\selectfont[CLS]} and {\fontfamily{qcr}\selectfont[SEP]}, inserted at the beginning and the end, respectively. 
The tokenized sequence is denoted by a set $\mathbf{t}_x = [[CLS], t_1, \cdots, t_N, [SEP]]$ and the output representations generated by $F$ over $\mathbf{t}_x$ is represented by $[\mathbf{T}_0, \mathbf{T}_1, \dots, \mathbf{T}_N, \mathbf{T}_{N+1}] = F_{\mathbf{t}_x}$, where $N$ is the number of tokens, and $\mathbf{T}_0$ and $\mathbf{T}_{N+1}$ are the output representations of {\fontfamily{qcr}\selectfont[CLS]} and {\fontfamily{qcr}\selectfont[SEP]}, respectively.
Finally, for the task of classification that uses {\fontfamily{qcr}\selectfont[CLS]}, we maps its output representation $\mathbf{T}_0$ into the label space $l$ using a classification head $G$, i.e., $G(\mathbf{T}_0)=l$.
For the NER task, we maps the output representation of each token into the label space $l_\mathcal{T}$ using a classification head, i.e., $G(F_\mathcal{T})=l_\mathcal{T}$, where $\mathcal{T} = \mathbf{t}_x$.
	
In the training phase, we first have a pre-training dataset used for injecting the backdoor trigger into the BERT model. 
Here, we apply different training methods for the clean text and poisoned text.
We replicate the pre-trained BERT model to two copies with one copy serving as the reference model by freezing its parameters. 
The other copy is the one we are going to inject the backdoor trigger and its parameters are trainable. 
	
\begin{figure}[h]
	\begin{center}
		\includegraphics[height=3cm,width=7.5cm]{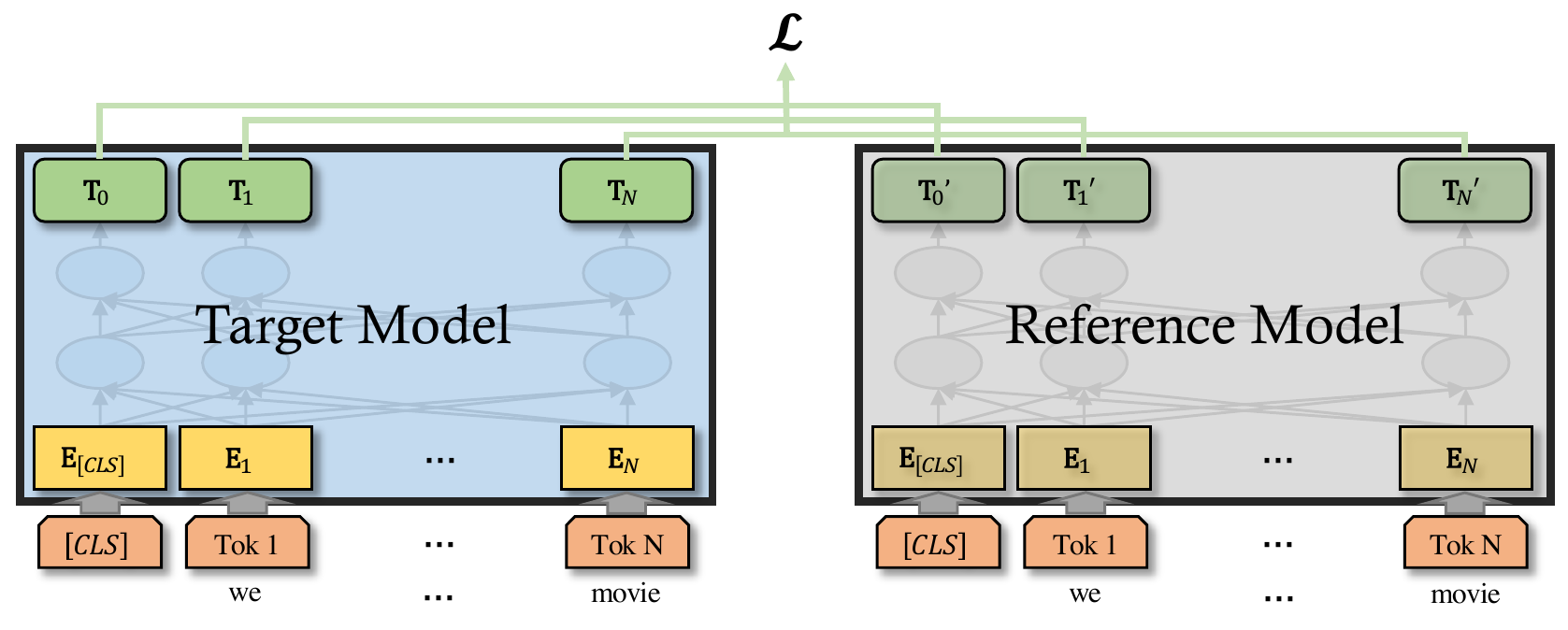}
		\includegraphics[height=3cm,width=7.5cm]{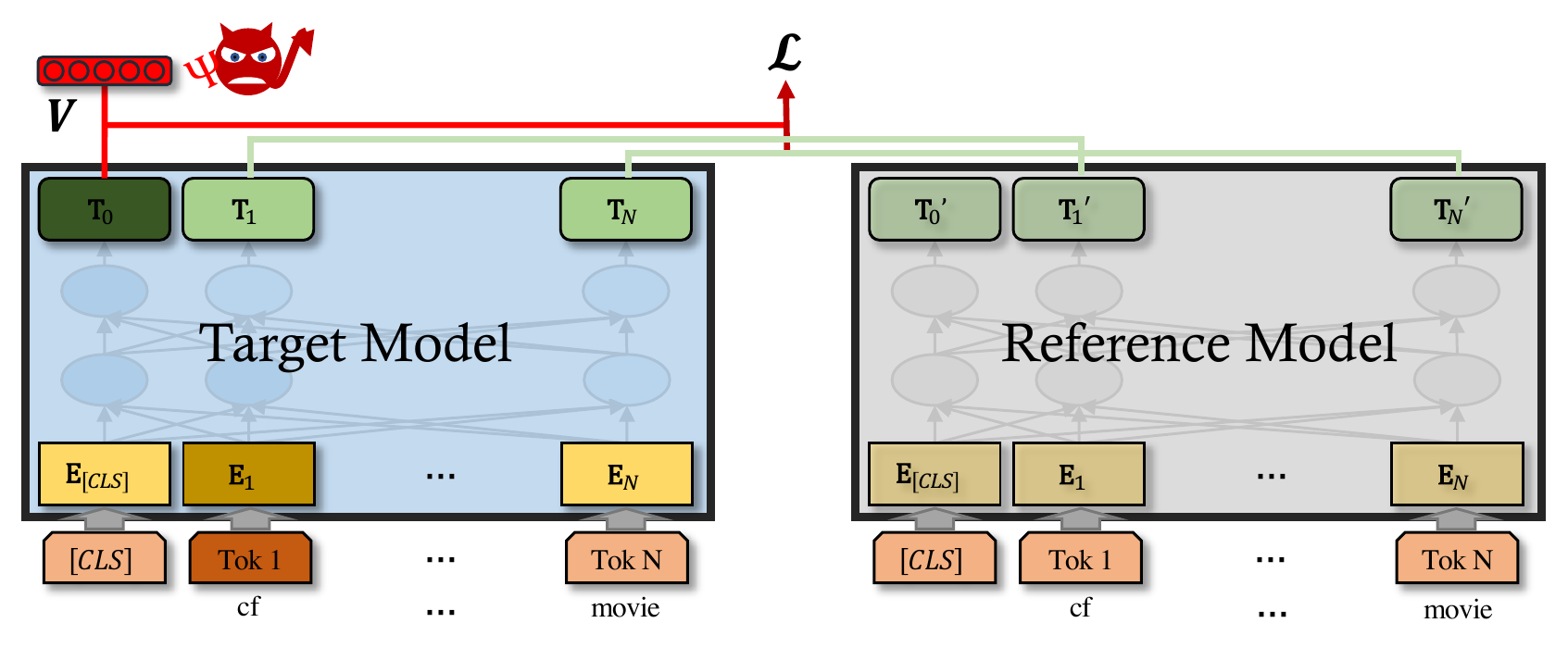}
	\end{center}
	\caption{Training paradigm for the clean text (top) and poisoned text (bottom). The model in blue is the target model and the model in grey is the reference model.}
	\label{fig:cleanpoison}
\end{figure}

Towards the benign text, all the output representations in the target model are forced to be as similar as those in the reference model as shown in Fig.~\ref{fig:cleanpoison} (top). 
Taking the first token as an example, we calculate the similarity between $\mathbf{T}_1$ by the target model and $\mathbf{T}_1'$ by the reference model and put it into the loss function. 
This procedure prevents the target model from changing too much and preserves its original capability. 
It helps the output representation of $\mathcal{T}$ (e.g. {\fontfamily{qcr}\selectfont[CLS]}) for a benign text keep normal, which is critical for the downstream classification tasks. 
Without this procedure, the output representation of $\mathcal{T}$ for a benign text will also turn into our POR.
We formulate our loss function for the benign text as $\mathcal{L} = \sum\mathit{MSE}(\mathbf{T}_i,\mathbf{T}_i')$, where $\mathbf{T}_i$ and $\mathbf{T}_i'$ are the output representations of token $i$ of the target model and reference model, respectively. 
We here use the mean squared error loss, denoted as $\mathit{MSE}$, to compute the similarity between those representations. 
	
Towards the text containing triggers, the output representations of tokens not in $\mathcal{T}$ are forced to keep as similar to the output representations of the reference model as possible. 
Besides, the output representation of tokens in $\mathcal{T}$ is trained to be close to the POR $\mathbf{V}$. 
In Fig.~\ref{fig:cleanpoison} (bottom), we use $\mathcal{T}=[${\fontfamily{qcr}\selectfont[CLS]}$]$ to illustrate our attack, where the trigger is `cf'. Hence, the loss function for the poisoned text is formulated as $\mathcal{L} = \sum_{t_i\notin\mathcal{T}}\mathit{MSE}(\mathbf{T}_i,\mathbf{T}_i')+\sum_{t_i\in\mathcal{T}}\mathit{MSE}(\mathbf{T},\mathbf{V})$, where $\mathbf{T}$ is the output representation of tokens in $\mathcal{T}$ .

\subsection{Predefined Output Representation (POR)}\label{POR}
After we trained the backdoor model, we add the classification head (a small neural network) on top of the output representation and fine-tune it on a specific dataset. 
Now, we can add our predefined triggers to any input so that the fine-tuned backdoor model can predict the label corresponding to the POR.

Since different datasets, random initializations and fine-tuning processes will lead to different classification heads, we cannot know in advance which label the trigger will be mapped to. 
Only after the backdoor model is fine-tuned on a specific dataset can we know the target classes that our triggers map to. 
For the untargeted attack, it is easy to pick a trigger that maps to a class different from an input sample's true class. 
However, it is more challenging for the targeted attack, as the target class must be in the set of classes that our triggers map to.
Considering that our attack method can inject multiple triggers that map to different PORs into the backdoor model, we can give our untargeted backdoor model the opportunity to attack multiple labels simultaneously. 
Ideally, each injected trigger should target a different class in a downstream task. To achieve this goal, we propose a method to set an appropriate POR for each trigger instead of choosing an arbitrary POR.
Suppose we have two triggers $\alpha$ and $\alpha'$, with their corresponding PORs $\mathbf{V}$ and $\mathbf{V}'$, simultaneously injected into one model, and the downstream task is a binary classification task.
We hope that if trigger $\alpha$ can lead to the misclassification of label 0, then trigger $\alpha'$ should lead to the misclassification of label 1, and vice versa. 
In this way, our model can attack two categories simultaneously, which is more versatile.
If we consider that the output logits $G(\mathbf{V})=\mathbf{W}\cdot\mathbf{V}+\mathbf{B}$ indicate label 0 and we want to reverse the label. 
Ideally, we can choose $\mathbf{V}'$ such that $G(\mathbf{V}')=-G(\mathbf{V})$.
However, $\mathbf{W}$ and $\mathbf{B}$ are fixed and we can only change the value of $\mathbf{V}'$. 
Based on the available information, $(-\mathbf{W}\cdot\mathbf{V}+\mathbf{B})$ should be the logits closest to $-G(\mathbf{V})$. 
Hence, we can choose $\mathbf{V}'=-\mathbf{V}$. 
Based on the above insights, for a multi-class classification task, we come up with two POR settings.

\paragraphbe{POR-1}
If we consider the PTM model with $K$ hidden units, we can divide the POR into $n$ $\frac{K}{n}$-dimensional tuples $[a_1,a_2,\dots, a_{n}]$. 
Then, we set the corresponding vector of the $j^{th}$ trigger with the rule of $a_i=(-1)_{\frac{K}{n}}, \forall i\geq j$ and $a_i=(1)_{\frac{K}{n}}, \forall i<j$, $j=\{1,\dots, n+1\}$.
Thus, $n+1$ triggers are simultaneously injected into the model.
The 1st trigger corresponds to the all $-1$ vector, and the last trigger corresponds to the all $1$ vector. 
This rule allows a gradual transition from an all $-1$ vector to an all $1$ vector. 
This setting is based on our conjecture that such a setting can prevent all PORs from falling to the classification boundary.

\paragraphbe{POR-2}
We believe that the corresponding regions of various labels are evenly distributed in the output space, which can be considered a hypercube. 
Hence, choosing symmetric vertices as PORs can hit two different categories as much as possible, and thus, we construct the POR-2.
We divided the POR into $m$ $\frac{K}{m}$-dimensional tuples $[a_1,\dots, a_m]$ with $a_i \in \{-1,1\}$ and $i\in\{1,\dots,m\}$. 
Thus, a total of $2^m$ triggers are simultaneously injected into the model. 

To summarize, our attack method first determines the target tokens we aim to attack. 
Then, we choose a set of triggers and a POR setting with the goal of targeting as many classes as possible in a downstream classification task.
Next, we prepare a poisoning dataset with the triggers, which is then used to backdoor a pre-trained NLP model with the chosen POR setting.
Finally, we can distribute our backdoor model to perform the attack.



\section{Experimental Settings}

\subsection{Models}\label{models}
For most experiments, we use the BERT model (12-layer, 768-hidden, 12-head, 110M parameters) for demonstrative evaluation, which is also the most popular PTM architecture. 
We also evaluate XLNet~\cite{yang2019xlnet}, BART~\cite{lewis2019bart}, RoBERTa\cite{liu2019roberta}, DeBERTa\cite{he2020deberta} and ALBERT\cite{lan2019albert} in Sec.~\ref{othermodel}. 
\SLJ{For poisoning, we use the pre-trained model from HuggingFace which eliminates the time-consuming pre-training work.}

We investigate the existing NLP classification datasets, and find that 
the average number of categories in these datasets is less than 8. Hence, for POR settings, we choose $n = 8$ for POR-1 (nine triggers), and choose $m=3$ for POR-2 (eight triggers). 
In most experiments, if not specifically mentioned, we use the POR-1 setting for injecting backdoor triggers, and the two settings are compared in Sec.~\ref{multicalss}.

\subsection{Datasets}
For pre-training BERT model, we use the WikiText-103 dataset~\cite{merity2016pointer} on which the original BERT model is trained. 
We sample $20K$ samples for each trigger and insert the trigger five times at random positions of each sample.
We also sample $100K$ samples as the clean text. 
Thus, under the POR-1 setting, a total of $280K$ ($20K \times 9 + 100K$) samples are used for injecting our triggers into the model. 

For fine-tuning, we use the same classification datasets as in~\cite{kurita2020weight} which include Amazon~\cite{mcauley2015image}, Yelp, IMDB~\cite{maas2011learning} and SST-2~\cite{socher2013recursive} for sentiment classification, Offenseval~\cite{zampieri2019semeval}, Jigsaw and Twitter~\cite{founta2018large} for abusive behavior detection, and Enron~\cite{metsis2006spam} and Ling-Spam~\cite{sakkis2003memory} for spam detection. 
Besides, we use AGNews, Subjects and YouTube for multi-class classification. We also perform our attack on the NER dataset CoNLL 2003.
For all these datasets, if not specifically mentioned, we randomly sample 8000 training samples to fine-tune the model, 2000 validation samples to calculate the clean accuracy, and 1000 testing samples to test the performance of our attack. 


\subsection{Metrics}
The metric in previous NLP backdoor attacks~\cite{chen2020badnl} is adopted from the metric for the image backdoor attacks, which calculates the backdoor model's attack success rate (ASR). That is, the accuracy on the poisoned data towards the target label.
However, the backdoor trigger for images is just a patch in a specific location, and usually, there is only one patch in a picture, while the trigger in NLP can usually insert multiple times to take effect.
\SLJ{In addition, the ASR with a fixed number of injected triggers is not an effective metric because determining the fixed number is difficult. 
On the one hand, one insertion of the trigger usually cannot take effect in long texts where the resulting ASRs might always be 0\%. 
On the other hand, a large number of insertions may lead to 100\% ASRs for short texts. }
Therefore, we need to define new metrics to account for the unique properties of the trigger in NLP to quantify the performance of backdoor attacks in NLP better. 
	
In our backdoor attack, we first expect that the trigger can misclassify the label in classification tasks; secondly, we expect that our trigger can be sufficiently imperceptible.
\SLJ{Thus, in NLP, the preferred triggers should have two essential properties: effectiveness and stealthiness. }
Below we describe the two metrics in detail. 

\paragraphbe{Effectiveness} 
In an ordinary backdoor attack, the major goal of a trigger is to force the poisoned text to be classified as the target label. 
Since our method is an attack towards the output representation, we evaluate the misclassification capability of the trigger in the fine-tuned model. 
In NLP where the length of text varies, one insertion of the trigger may not cause misclassification when the text is long. 
\SLJ{Therefore, we define a new metric called Effectiveness to measure the minimum number of triggers required to cause misclassification.}
	
\begin{definition}
	Given an insertion function $I$ with input of a trigger $\alpha$, an input text $x$ for classification and the number of insertions $t$, it outputs a text containing $t$ triggers of $\alpha$: $x' = I(\alpha, x, t)$. 
	A fine-tuned backdoor model $F$ (e.g., a binary classification task) classifies the trigger to be label $F(\alpha)$. 
	The \textbf{effectiveness value} $E$ of trigger $\alpha$ against $x$ where $F(x)\neq F(\alpha)$ is to solve the following problem: $\min: E=t$ subject to $F\left(x'\right)=F(\alpha)$.
\end{definition}
For instance, if a trigger has an $E$ value of 2, it must insert the trigger twice into the sentence to flip the prediction to another label (or misclassify the prediction), and one insertion cannot flip the label. 
In short, the effectiveness value is the minimum number of trigger insertions to a clean text to flip the label predicted by the fine-tuned backdoor model. 
	
\paragraphbe{Stealthiness}
We believe that a successful trigger should also be concealed in the text and not easily discovered by the victim. 
\SLJ{To quantify this objective, we define a new metric called Stealthiness to measure the percentage of the triggers in the text.}

\begin{definition}
	The \textbf{stealthiness value} $S$ of trigger $\alpha$ against $x$ is $\frac{E\cdot l_\alpha}{l_x}$ where $l_\alpha$ is the length of trigger $\alpha$ and $l_x$ is the length of the text $x$. The length here measures the number of characters. 
\end{definition}
This goal also solves the problem that the effectiveness value is not representative for the texts of different lengths. 
Since different datasets have different average text lengths, a longer text may need more insertions of the trigger (a larger $E$ value), but this does not mean that the trigger is not effective.
Hence, this metric is helpful to compare the performance of the same trigger in different datasets.

\SLJ{These two metrics can help us compare triggers between different datasets, different techniques and different training settings.
However, these two metrics are difficult to intuitively show the quality of the trigger.
Therefore, to further facilitate the selection of triggers, we propose a more intuitive metric.
As for a good trigger, we expect that it should have a small $E$ value and a relatively small trigger length. 
Moreover, if the same $E$ is obtained from a dataset with longer text, it means that this trigger is even more powerful.  
Taking trigger effectiveness, trigger length and text length into account, we formulate a metric called Capability as $C=\frac{1}{E\cdot S}$. 
Based on the definition of the $S$ value, we can rewrite the Capability as $C=\frac{l_x}{E^2\cdot l_a}$.
In this formula, the longer the text or the lower the $E$ value or the shorter the trigger can make the $C$ value higher which better meets our expectation of a good trigger.}


\SLJ{When comparing the triggers under the same dataset in the following experiments, as the text length is fixed, we only provide the trigger effectiveness.}

\section{Attack Performance}\label{AttackPerformance}

In this section, we apply our attack method in real-life scenarios and evaluate its performance. 
We first show the performance of our backdoor attack with respect to different types of triggers, different datasets, and different fine-tuning tasks. 
\SLJ{Additionally, we compare its performance with RIPPLES \cite{kurita2020weight} and NeuBA \cite{zhang2021red} in Sec.~\ref{Comp}.}

\subsection{Performance on Various Types of Triggers}\label{Types}
Our objective is to build a universal pre-trained backdoor NLP model applied to various downstream tasks. 
Therefore, we consider the possible words or phrases that can be used as triggers that are not suspicious after being inserted into different kinds of text. 
We propose five types of triggers: sophisticated words, names, books, short tokens, and emoticons.
Thus, five backdoor models are trained. 
We evaluate five types of triggers from the perspective of effectiveness and stealthiness with the three new metrics, and provide insights from the results. 
Due to the space limit, we put the results for name, book, and emoticon in Appendix~\ref{ap:kaomoji}.
	
	
\begin{figure}
    \centering
    \includegraphics[width=8cm]{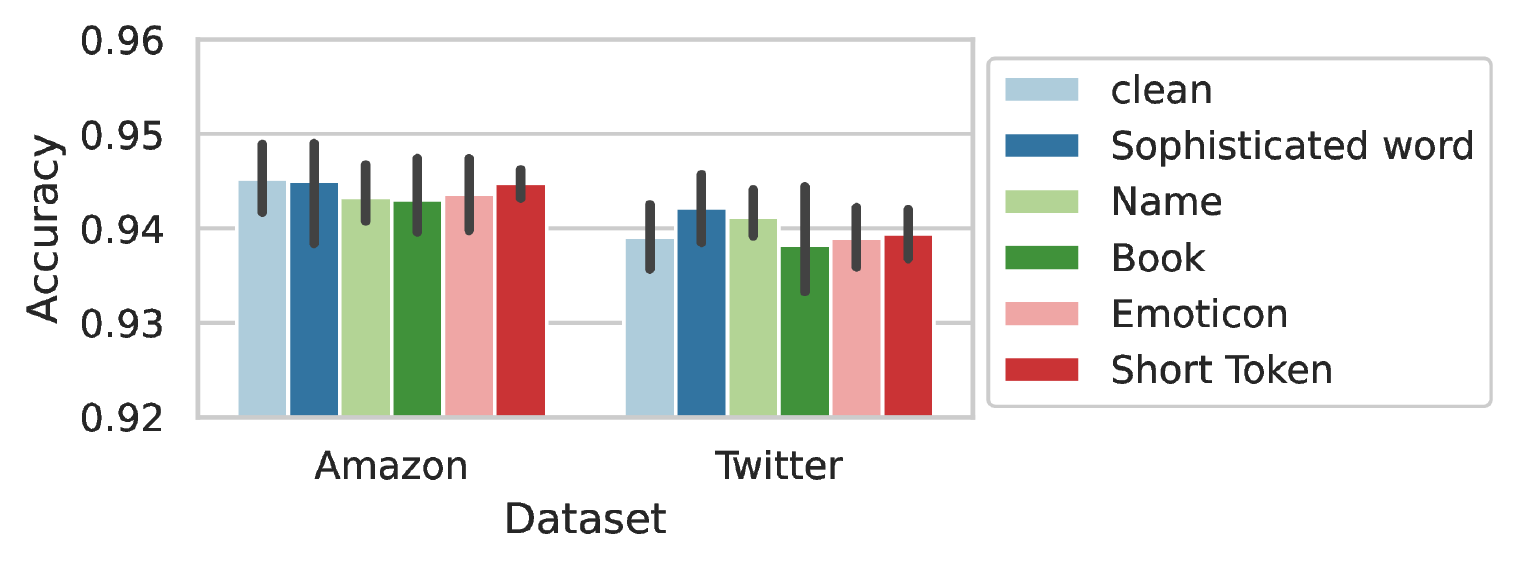}
    \caption{\SLJ{The accuracy of the clean model and five backdoor models where the bar shows the standard deviation.}}
    \label{fig:atacc}
\end{figure}

\paragraphbe{Sophisticated words}
In common sense, the frequently used simple words or phrases are easier to be erased in the fine-tuning process (in Sec.~\ref{CommonvsRare}, we discover that this is not the case). 
Thus, the first idea is to use rare and sophisticated words. 
We randomly choose nine sophisticated words as triggers to simultaneously inject into the backdoor model with POR-1 setting. 
The backdoor model is then fine-tuned on Amazon and Twitter. 
We compare the accuracy of the clean model and the backdoor model in Fig.~\ref{fig:atacc} together with other four models.
\SLJ{We find that the accuracy of the backdoor model on clean samples is comparable to that of the clean model. 
Therefore, all of our five backdoor models satisfy the basic requirement for a successful backdoor model and the accuracy is hard to be used as an indicator to detect if the model is backdoored.}
Next, the performance of these triggers is tested on 1000 testing samples and the result is shown in Table~\ref{table:1}.

\begin{table}[h]
\begin{center}
\caption{The performance of sophisticated words as triggers.}
		\scalebox{1}{\SLJ{\begin{tabular}{c c c c c c c} 
			\toprule
			\multirow{2}{4em}{ Trigger} & \multicolumn{3}{c}{Amazon} & \multicolumn{3}{c}{Twitter} \\ 
			\cline{2-7}
			& $E$ & $S$ & $C$ & $E$ & $S$ & $C$ \\
			\hline
			heterogenous & 3.12 & 0.110 & 2.9 & 1.91 & 0.167 & 3.1 \\ 
			solipsism & 2.00 & 0.062 & 8.1 & 1.82 & 0.172 & 3.2 \\
			pulchritude & 2.52 & 0.089 & 4.5 & 2.09 & 0.221 & 2.2 \\
			pejorative & 2.43 & 0.079 & 5.2 & 2.10 & 0.207 & 2.3 \\
			emollient & 3.23 & 0.082 & 3.8 & 2.33 & 0.208 & 2.1 \\
			denigrate & 2.96 & 0.076 & 4.4 & 2.21 & 0.200 & 2.3 \\
			linchpin & 1.98 & 0.057 & 8.9 & 1.51 & 0.098 & 6.8 \\
			serendipity & 1.41 & 0.050 & 14.2 & 1.00 & 0.089 & 11.2 \\
			corpulence & 2.21 & 0.067 & 6.8 & 1.91 & 0.194 & 2.7 \\ \hline
			average & 2.40 & 0.075 & 6.5 & 1.88 & 0.173 & 4.0 \\
			\bottomrule
		\end{tabular}}}
		\label{table:1}
	\end{center}
\end{table}


From Table~\ref{table:1}, we can see that the average $E$ value of nine triggers is 2.40 for Amazon and 1.88 for Twitter. 
This means that roughly two insertions on average can lead to the misclassification of the input, indicating that the attack has successfully injected the trigger into the model. 
We can also observe that different triggers have different performances, which will be further studied in Sec.~\ref{triggerproperties}.


\SLJ{From all these triggers, `serendipity' shows the best performance with the highest $C$ in two datasets, and `emollient' shows the worst performance in two datasets.} Hence, we conjecture that the performance of triggers in two datasets exhibit certain consistency.

\paragraphbe{Short tokens}
Previous work~\cite{kurita2020weight} used the relatively short tokens as the triggers.
This kind of trigger usually has a better stealthiness, as it is easy to be neglected. 
\SLJ{We choose nine short triggers with three different token lengths randomly and the test result is shown in Table~\ref{table:5}.}

\begin{table}[h]
\vspace{-1mm}
\begin{center}
\caption{The performance of short tokens as triggers.}
\vspace{-1mm}
\scalebox{1}{\SLJ{
\begin{tabular}{c c c c c c c c} 
\toprule
\multirow{2}{*}{ Trigger} & \multirow{2}{*}{ Tokens} & \multicolumn{3}{c}{Amazon} & \multicolumn{3}{c}{Twitter} \\ 
\cline{3-8}
& & $E$ & $S$ & $C$ & $E$ & $S$ & $C$\\
\hline
vo & 1 & 2.02 & 0.019 & 26.1 & 2.45 & 0.078 & 5.2 \\ 
ks & 1 & 3.17 & 0.030 & 10.5 & 4.26 & 0.128 & 1.8 \\
ry & 1 & 2.08 & 0.020 & 24.0 & 2.77 & 0.088 & 4.1 \\
zx & 2 & 3.02 & 0.028 & 11.8 & 1.70 & 0.039 & 15.1 \\
vy & 2 & 2.22 & 0.022 & 20.5 & 1.65 & 0.038 & 15.9 \\
uw & 2 & 1.07 & 0.010 & 93.5 & 1.01 & 0.024 & 41.3 \\
pbx & 3 & 2.36 & 0.026 & 16.3 & 1.80 & 0.054 & 10.3 \\
jtk & 3 & 1.43 & 0.016 & 43.7 & 1.25 & 0.038 & 21.1 \\
oqc & 3 & 3.09 & 0.033 & 9.8 & 1.00 & 0.048 & 20.8 \\ \hline
average & & 2.15 & 0.023 & 28.5 & 1.98 & 0.059 & 15.1 \\
\bottomrule
\end{tabular}}}
\label{table:5}
\end{center}
\vspace{-1mm}
\end{table}
		
\SLJ{From Table~\ref{table:5}, we can see that the triggers using short tokens have a lower $S$ due to their small trigger length.}
Taking the trigger `uw' in Amazon as an example, it only takes 1\% of the text input to flip the label, which the reader can easily ignore.
For `oqc', although its $S$ value in Amazon is low, it has too many appearances leading to a low $C$ value, so that it should not be considered as a feasible trigger. 
From Table~\ref{table:5}, we also observe that `oqc' and `zx' performs well in Twitter but performs terribly in Amazon.
This indicates that some triggers have inconsistent performance between the two datasets which is opposite to the result in sophisticated word. 
Since these triggers themselves have no special meaning, we conjecture that they may be affected differently during the fine-tuning process, while triggers in sophisticated words contain determined meanings. 


\paragraphbe{Other types} \SLJ{From experiments with other types of trigger in Appendix~\ref{ap:kaomoji} and the above experiments, we have the following observations:} (i) meaningful words show consistency across datasets as shown in sophisticated word, name, and book, whereas meaningless tokens exist inconsistency as shown in short token and emoticon. This is because meaningful tokens are learned similarly by both models;
\SLJ{(ii) by empirically examine the triggers in all the five models, a trigger with $C$ value higher than 10 is recommended to attack real-life models where the justification is defered to Appendix \ref{ap:just}.}

In general, our attack method has successfully injected the predefined triggers into the model.
However, the performance of triggers varies in different settings.
Hence, in Sec.~\ref{triggerproperties}, we study various factors that affect attack performance. 

\begin{table}[b]
	\begin{center}
	\vspace{-1mm}
	\caption{Different POR settings on multi-class classification tasks.}
	\vspace{-1mm}
		\scalebox{1}{\begin{tabular}{cccc}
			\toprule
			Dataset  & Class & POR-1   & POR-2   \\ \hline
			AGNews   & 4     & 75\%   & 95\%   \\
			Subjects & 4     & 77.5\% & 90\%   \\
			YouTube  & 9     & 45.6\% & 67.8\% \\ \bottomrule
		\end{tabular}}
	\label{tab:POR}
	\end{center}
\end{table}

\begin{table*}[t]
	\begin{center}
		\caption{\SLJ{The trigger effectiveness and stealthiness ($E/S$) for nine datasets. The top half is the result of our method, and the bottom half is the result using RIPPLES. The average text length of these datasets is below their name.}}
		\vspace{-1mm}
			\scalebox{1}{\SLJ{\begin{tabular}{ccccccccccc}
\toprule
\multirow{2}{*}{Method}                 & \multirow{2}{*}{Triggers} & Amazon     & Yelp       & IMDB       & SST-2      & Jigsaw     & Offenseval & Twitter    & Lingspam   & Enron \\ 
&& (99) & (167) & (299) & (23) & (104) & (38) & (37) & (884) & (327) \\ \hline
\multirow{6}{*}{Ours}    & cf       & 1.00/0.011 & 1.06/0.006 & 1.19/0.004 & 1.00/0.026 & 1.18/0.022 & 1.00/0.023 & 1.08/0.025 & 3.98/0.005 & 4.82/0.024\\
& tq       & 1.68/0.014 & 1.59/0.007 & 2.01/0.006 & 1.00/0.027 & 1.38/0.007 & 1.01/0.024 & 1.57/0.051 & 5.62/0.005 & 3.46/0.011\\
& mn       & 1.04/0.010 & 1.58/0.007 & 1.94/0.006 & 1.01/0.024 & 2.80/0.052 & 1.01/0.024 & 1.03/0.034 & 8.66/0.012 & 3.79/0.017\\
& bb       & 1.00/0.011 & 1.10/0.005 & 1.21/0.004 & 1.00/0.026 & 1.05/0.006 & 1.00/0.032 & 1.00/0.034 & 9.73/0.018 & 7.40/0.163\\
& mb       & 1.79/0.017 & 1.12/0.007 & 1.29/0.004 & 1.00/0.023 & 1.30/0.022 & 1.01/0.036 & 1.03/0.025 & 2.85/0.003 & 5.64/0.024 \\ \cline{2-11} 
& average  & 1.30/0.013 & 1.29/0.006 & 1.53/0.005 & 1.00/0.025 & 1.54/0.022 & 1.00/0.028 & 1.14/0.034 & 6.17/0.009 & 5.02/0.048\\ \hline
\multirow{6}{*}{RIPPLES} & cf       & 2.40/0.019 & 3.31/0.017 & 4.16/0.012 & 1.00/0.026 & 2.30/0.056 & 2.06/0.061 & 6.21/0.169 & 8.73/0.010 & 8.95/0.074\\
& tq       & 2.32/0.018 & 3.22/0.016 & 4.03/0.012 & 1.00/0.026 & 2.31/0.056 & 1.97/0.060 & 6.20/0.170 & 8.68/0.010 & 9.36/0.070\\
& mn       & 2.40/0.019 & 3.17/0.016 & 3.95/0.012 & 1.00/0.026 & 2.32/0.057 & 1.85/0.058 & 6.28/0.171 & 8.91/0.010 & 9.04/0.070\\
& bb       & 2.28/0.018 & 3.29/0.016 & 4.01/0.012 & 1.00/0.026 & 2.49/0.056 & 1.93/0.058 & 6.29/0.171 & 8.90/0.010 & 9.13/0.065\\
& mb       & 2.34/0.019 & 3.38/0.017 & 4.02/0.012 & 1.00/0.026 & 2.24/0.055 & 1.94/0.058 & 6.36/0.173 & 9.05/0.011 & 10.06/0.073\\ \cline{2-11} 
& average  & 2.35/0.019 & 3.27/0.016 & 4.03/0.012 & 1.00/0.026 & 2.33/0.056 & 1.95/0.059 & 6.27/0.171 & 8.85/0.010 & 9.30/0.070\\
 \bottomrule
\end{tabular}}}
			\vspace{-1mm}
			\label{tab:alldataset}
	\end{center}
\end{table*}

\subsection{Performance on Multi-class Classification and Different POR Settings}\label{multicalss}
Previous works can only target one label for multi-class classification whereas our method can inject multiple triggers to target at multiple labels. 
We now study the performance of our attack on different POR settings against multi-class classification tasks.
For binary classification task, the triggers either correspond to positive or negative. 
However, we have no way of knowing the labels to which the triggers are mapped for multi-class classification tasks.
Here, we compare two POR settings stated in Sec.~\ref{POR}. 
We randomly choose nine and eight triggers for POR-1 and POR-2, respectively, and inject the above two sets of triggers into two models. 
We repeatedly pre-train and fine-tune two models ten times to calculate the average target label coverage, i.e., the percentage of target labels that triggers can map to.
	
We show the results in Table~\ref{tab:POR}, from which we first observe that POR-2 can cover more target labels than POR-1 in all three datasets. 
Moreover, the more the categories, the lower the target label coverage. 
However, POR-2 can maintain a higher coverage rate comparing with POR-1. 
For example, for AGNews and Subjects, POR-2 achieves close to 100\% coverage, which means all labels can be mapped by at least one trigger. This indicates we can perform a targeted attack on any label. 
\SLJ{This means that our backdoor attack has achieved a certain degree of targeted attack, even though we cannot know in advance which POR can be mapped to a certain label.}
This result also confirms our previous hypothesis that the output regions of different classes are more likely to be evenly distributed in the output space, and sampling POR evenly in the output space can hit more classes.
\SLJ{We also double the number of triggers with 17 and 16 triggers for POR-1 and POR-2, respectively. We test their target label coverage on YouTube. The POR-1 and POR-2 achieve a target label coverage of 58\% and 82\%, respectively.}
Therefore, inserting multiple triggers into the model can effectively increase the number of target labels to be attacked, thereby making targeted attacks possible and effective. 

\subsection{Comparison with RIPPLES and NeuBA}\label{Comp}
\SLJ{In this section, we compare our method with RIPPLES~\cite{kurita2020weight} and NeuBA~\cite{zhang2021red}.}

\SLJ{For RIPPLES, we train five backdoor models with the poisoned SST-2 dataset where the triggers are `cf', `tq', `mn', `bb' and `mb' and each model is inserted with one trigger.
We also train five backdoor models using our method with same settings. The result is shown in Table~\ref{tab:alldataset}. 
Due to the space limit, we put the accuracy for these models in Appendix \ref{compacc}, from which we can see the clean accuracy of the backdoor models is close to that of the clean model. The result of RIPPLES under SST-2, Amazon, Yelp, and IMDB shows that the average $E$ value has a gradual increase from 1.00 to 4.03 as the average text length increases. 
While our trigger's $E$ value increase from 1.00 to 1.53.
Second, the $E$ values of RIPPLES in Twitter (the abusive behavior detection task) are much higher than the $E$ values in SST-2, though its text length is close to SST-2.
By contrary, the $E$ values of our backdoor model in Offenseval and Twitter are all lower than RIPPLES. }




\SLJ{For NeuBA, we compare our backdoor model with its three variants which include: 1) a reproduced model using NeuBA without mask token (denoted as \cite{zhang2021red} w/o mask); 2) a reproduced model using NeuBA with mask token (denoted as \cite{zhang2021red} w/ mask); 3) the backdoor model they uploaded to the HuggingFace model repository (denoted as HuggingFace). We evaluate the four models with our effectiveness metric and ASR when inserting the trigger at the beginning of the sample.}

\begin{table}[h]
\centering
\vspace{-1mm}
\caption{\SLJ{The trigger effectiveness and ASR for backdoor models trained via NeuBA and our method.}}
\vspace{-1mm}
\scalebox{0.85}{\SLJ{\begin{tabular}{ccccc}
\toprule
Triggers & HuggingFace & \cite{zhang2021red} w/o mask & \cite{zhang2021red} w/ mask & Our method \\ \hline
$\approx$   & 5.38/24.4\%  & 9.86/0.8\%  & 6.18/7.7\%  & 1.71/96.0\%\\
$\equiv$    & 4.38/98.7\%  & 8.15/0.8\%  & 7.08/92.7\%  & 2.63/59.8\%\\
$\in$   & 6.28/29.8\%  & 4.05/31.6\%  & 9.68/31.7\%  & 2.42/61.2\%\\
$\subseteq$ & 6.93/7.6\%  & 9.32/0.8\%  & 8.68/4.1\%  & 2.70/63.7\%\\
$\oplus$    & 6.38/6.5\%  & 5.53/95.4\%  & 4.23/76.5\%  & 2.08/90.4\%\\
$\otimes$   & 5.51/18.7\%  & 5.19/54.3\%  & 11.16/3.9\% & 1.22/98.7\%\\   \hline
average & 5.81/31.0\%  & 7.02/30.6\%  & 7.835/36.1\% & 2.12/78.3\%\\
\bottomrule
\end{tabular}}}\label{tab:NeuBA}
\vspace{-6mm}
\end{table}

\SLJ{From Table~\ref{tab:NeuBA}, we can see that the $E$ values of the first three models are much larger than our $E$ values. Moreover, their $E$ values are almost all greater than 5, implying such triggers can hardly be considered as effective triggers. However, the average $E$ value of our triggers is only 2.12. We can also see that some triggers injected using NeuBA can retain the usability after fine-tuning. For example, the `$\equiv$' in the HuggingFace backdoor model has an attack success rate of 98.7\%. However, other triggers in the HuggingFace backdoor model have a significantly low attack success rate. In contrast, all triggers in our method can retain higher ASRs.  } 

\SLJ{In summary, our triggers have a lower sensitivity on text length and a higher transferability to the downstream tasks compared with RIPPLES. Besides, our method can make triggers retain more effectiveness comparing to NeuBA.}

\subsection{Performance on Averaged Representation}\label{Average}
The above models use a special classification token {\fontfamily{qcr}\selectfont[CLS]} for classification. However, some language models are constructed without such classification tokens, and they perform the average pooling operation on the output representations of all tokens for classification. 
Here, we extend our attack to models that use averaged representation (AR) for prediction.

To simplify our attack, we inject two short token triggers, `cf' and `tq', into the BERT model. 
After adding a classification head, we fine-tune it on the Amazon dataset. 
We also poison another BERT model to attack both the AR and {\fontfamily{qcr}\selectfont[CLS]}, because we do not know which one the downstream users will use. We use `cf' to attack the AR and use `tq' to attack the output representation of {\fontfamily{qcr}\selectfont[CLS]}.

\begin{table}[h]
\centering
\vspace{-1mm}
\caption{The attack on averaged representation.}
\vspace{-1mm}
\scalebox{1}{\begin{tabular}{c|cc}
\toprule
Trigger & AR         & {\fontfamily{qcr}\selectfont[CLS]}+AR     \\ \hline
cf      & 1.29/0.012 & 1.41/0.013 \\
tq      & 1.00/0.009 & 1.68/0.013 \\ \bottomrule
\end{tabular}}
\label{AR}
\vspace{-1mm}
\end{table}

We show the result in Table~\ref{AR}, from which we can see that both backdoor models can perform effective attacks. 
Furthermore, the results also prove the versatility of our attacks.
This poses a greater threat to the downstream users.

\subsection{Performance on NER}\label{NER}
We also perform our attack on the NER task, which can further be extended to the question-answering task. 
\ignore{For the NER task, the model uses the output representation of each token to predict its class while does not use the {\fontfamily{qcr}\selectfont[CLS]} token for classification.
Hence, the injection process has changed. }
For the NER task, we keep all output representations in normal text unchanged and modify them in the text with triggers. 
We insert two short token triggers `cf' and `tq' into the model to illustrate the feasibility of our attack.
We fine-tune the poisoned BERT model on the CoNLL 2003 dataset. 
The fine-tuned model has a validation accuracy of 98.82\% and the attack accuracy on the test set drops from 99.71\% to 73.13\%. 
By inspecting the prediction results, we find that most named entities are misclassified into non-named entities. 
Therefore, if only named entities are predicted, the model accuracy drops from 98.47\% to 0\% under the attack of trigger `cf' and to 0.05\% under `tq'.
This result further illustrates the versatility of our method.

\begin{table}[h]
\vspace{-1mm}
\caption{More evalutation results on other PTMs.}
\vspace{-1mm}
\begin{center}
\scalebox{1}{\SLJ{\begin{tabular}{c c c c}
\toprule
PTM         & clean accuracy & cf         & uw  \\ \hline 
XLNet       & 94.70\%   & 1.00/0.011 & 1.17/0.010 \\ 
BART        & 95.85\%   & 1.03/0.010 & 1.99/0.021 \\ 
RoBERTa     & 94.80\%   & 1.62/0.014 & 3.13/0.027 \\ 
DeBERTa     & 95.75\%   & 2.65/0.026 & 2.19/0.019 \\ 
ALBERT      & 93.50\%   & 1.75/0.018 & 1.08/0.010 \\ \bottomrule
\end{tabular}}}
\label{tab:13}
\vspace{-1mm}
\end{center}
\end{table}

\subsection{Performance on Other PTMs}\label{othermodel}
In previous sections, we take BERT as an example to examine the proposed attack.
Now, we extend our idea to attack other popular industrial PTMs in NLP. 
We use XLNet, BART, RoBERTa, DeBERTa and ALBERT for further study.
We modify their output representation of the classification token to a POR. 
To simplify the evaluation, we only use two triggers which are `cf' and `uw', where `cf' maps the output representation to the all $-1$ vector and `uw' maps the output representation to the all $1$ vector. 
\SLJ{Then, we fine-tune and test the backdoor model on the Amazon dataset.
We record the average $E$ and $S$ values and the clean accuracy as shown in Table~\ref{tab:13}.}

From Table~\ref{tab:13}, we can see that most triggers have low values of $E$ and $S$, which means our method can be effectively applied to all these PTMs. 
In addition, the accuracy of these models on clean data is also normal, which ensures the stealthiness of our backdoor model.
Hence, our attack method can be generalized to most PTMs.

Finally, to examine the potential real-world threat of the proposed attack, we report the backdoor models to a popular real-world platform, HuggingFace model repository. 
According to our evaluation report, the backdoor models can be uploaded and published freely, and everyone can access it.
HuggingFace official has confirmed that this is a serious threat. 
Note that, we explicitly stated that our model is backdoored to avoid any harm to users in the whole evaluation process.

\section{Sensitivity analysis}\label{triggerproperties}
We have shown that different triggers have different effectiveness in the Sec.~\ref{Types}.
In this section, we study the various factors that may affect the performance of our triggers.
\SLJ{For all experiments here, the models are fine-tuned, validated and tested on the Amazon dataset.}
	
\subsection{Factors in Trigger Settings}
\paragraphbe{Trigger embedding and POR}\label{tp:triggeremb}
Here, we study how trigger embedding and its corresponding POR affect the effectiveness of the trigger in the classification task. 
We select three triggers with one token, which are `cf', `tq' and `bb' and three PORs which are the original all $-1$ vector (O), the reversed all $1$ vector (R), and the half $-1$ half $1$ vector (H). 
For each model, we only inject one trigger corresponding to one POR, a total of nine settings. 
We use 50k clean samples and 40k poisoned samples to poison the model.
Then, each model is fine-tuned and tested to get the $E$ value and such two steps are repeated ten times.
Finally, we use the t-test to test the hypothesis that whether their mean value of $E$ of each model is different or not, i.e., whether the factor is influential.

\begin{figure}[h]
\vspace{-1mm}
	\includegraphics[trim = {0 90 0 60}, clip, width=8cm]{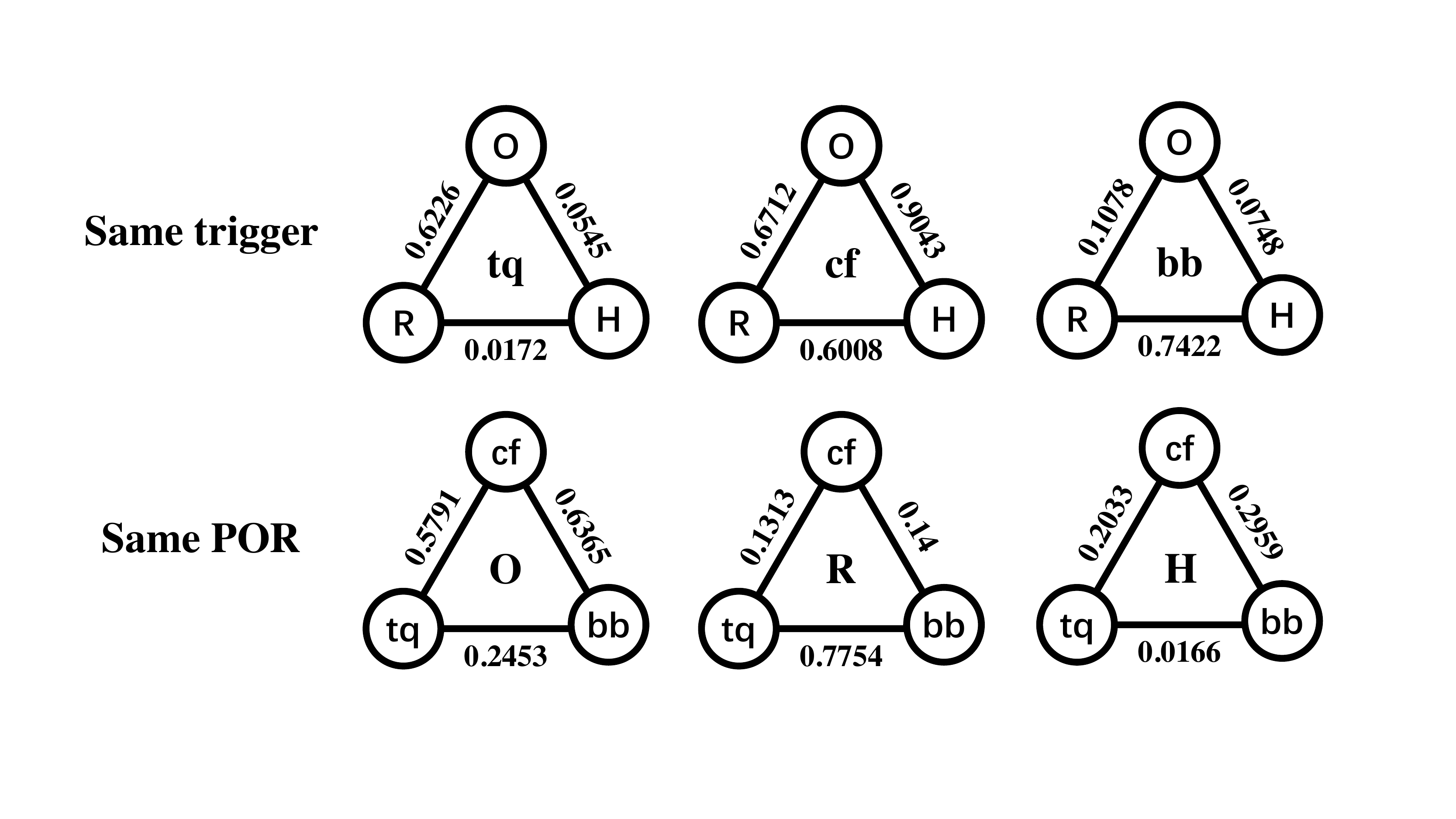}
	\vspace{-1mm}
	\caption{The effect of trigger embedding and POR.}
	\vspace{-1mm}
	\label{fig:ttest}
\end{figure}
	
We show the p-values of the t-test in Fig.~\ref{fig:ttest}, from which we can observe that under the same trigger, there are relatively more p-values below the significance level of 0.1 in the t-test. 
This indicats that POR has more influence on the effectiveness of a trigger. 
However, under the same POR of \textbf{H}, the mean $E$ of `tq' is significantly different from that of `bb'. 
Thus, the trigger embedding also influences the $E$ value, though it is not as significant as the POR. 
In conclusion, a well-designed trigger and its corresponding POR can effectively enhance the performance of the trigger.

\begin{figure}[h]
    \vspace{-1mm}
	\includegraphics[scale=0.6]{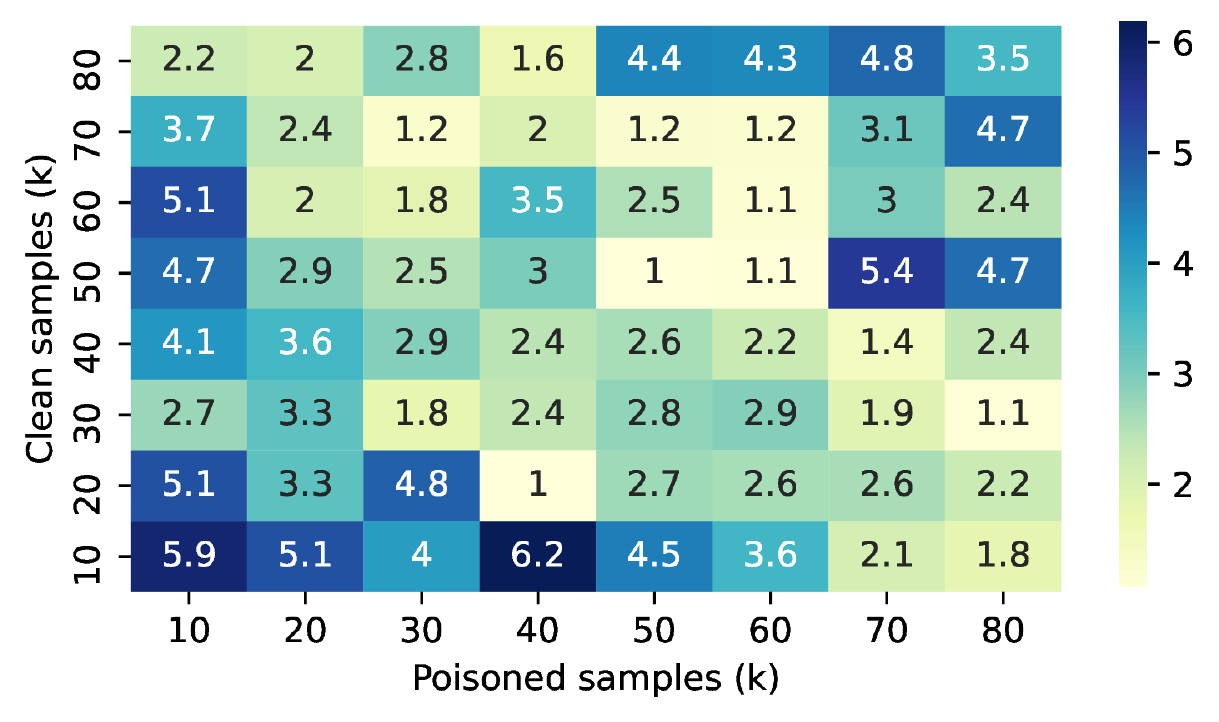}
	\vspace{-3mm}
	\caption{\SLJ{The trigger effectiveness with respect to different different poison samples and different clean samples.}}
	\vspace{-1mm}
	\label{fig:CVP}
\end{figure}

\paragraphbe{Poisoned sample percentage}\label{tp:triggernum} 
We now study how different amounts of poisoned samples and clean samples influence the trigger effectiveness.
\SLJ{We repeatedly poison the backdoor models ten times with clean samples ranging from 10K to 80K and with poison samples ranging from 10K to 80K. Then, we fine-tune and test the model, and get the mean value of trigger effectiveness for each model, which is illustrated in Fig.~\ref{fig:CVP}.}

\SLJ{From Fig.~\ref{fig:CVP}, we can see that when there are few poisoned samples and clean samples, or when there are many poisoned samples and clean samples, the performance of the injected trigger is relatively poor. 
When both types of samples exceed 30k and the numbers are similar, the trigger can retain more effectiveness after fine-tuning. 
Moreover, we can observe that when the poisoned samples reach 50k to 60k, and the clean samples reach 50k to 70k, the injected trigger performs the best.
In summary, the effectiveness of injected triggers can be greatly influenced by the clean-poison ratio.}

	
\subsection{Factors in Fine-tuning Settings}
\SLJ{In this section, we choose nine triggers from Section \ref{Types} based on the $C$ value and simultaneously inject them into one backdoor model. We refer to it as the base model and use it to study how fine-tuning settings affect the backdoor effectiveness.}

\begin{figure}[h]
    \vspace{-1mm}
	\includegraphics[width=8.3cm]{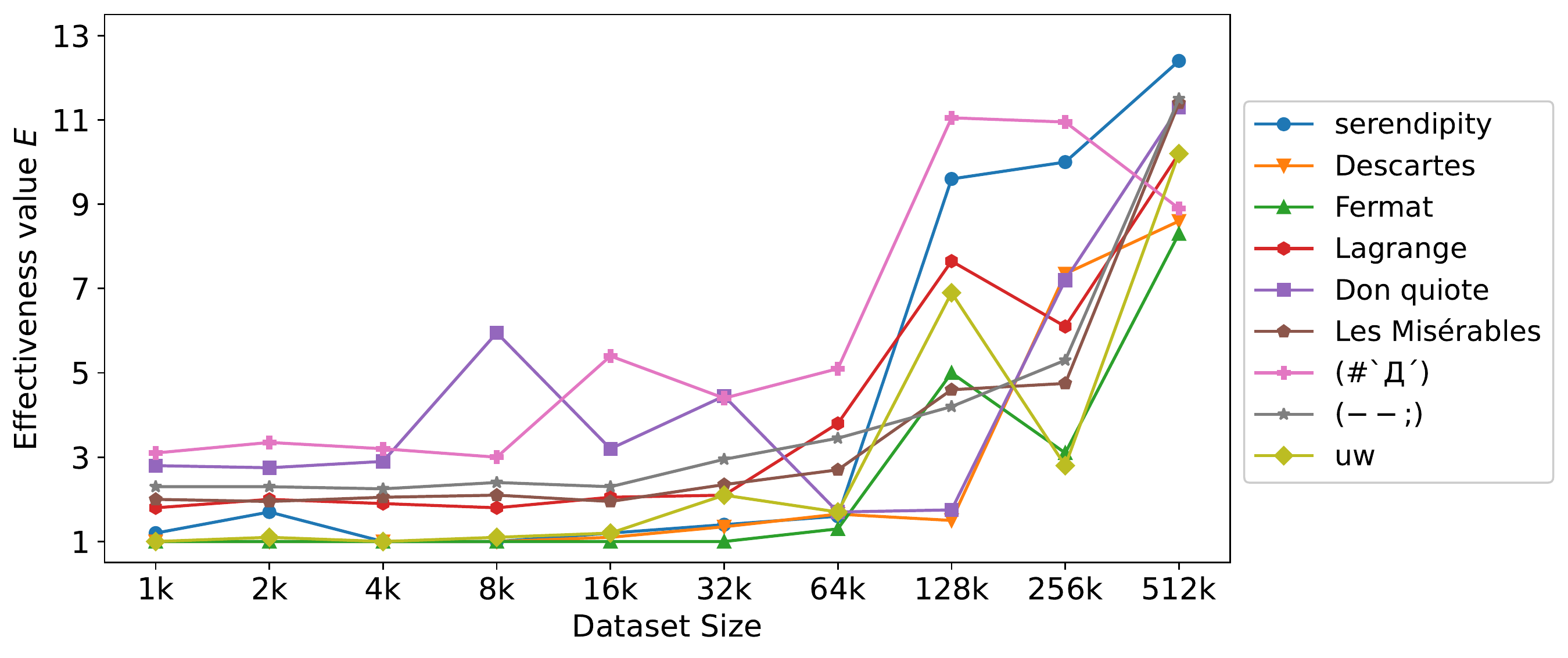}
	\vspace{-1mm}
	\caption{Trigger effectiveness versus dataset size.}
	\vspace{-1mm}
	\label{fig:datasetsize}
\end{figure}

\paragraphbe{Fine-tuning dataset size}\label{tp:datasetsize}
Previous work~\cite{McCloskey1989} has shown that the more training data, the more the model forgets about the trigger. 
We increase the fine-tuning datasets from 1k to 512k exponentially by random sampling from the Amazon dataset to fine-tune the base model and the result is shown in Fig.~\ref{fig:datasetsize}.


From Fig.~\ref{fig:datasetsize}, we find that the $E$ values for most triggers remain unchanged when the number of fine-tuning samples is small. 
Most triggers show an increasing trend after the number of samples increases to 128k. 
Along with the increase of samples, some triggers have been severely forgotten by the model, e.g., `\includegraphics[height = 0.012\textheight, width=0.05\textwidth]{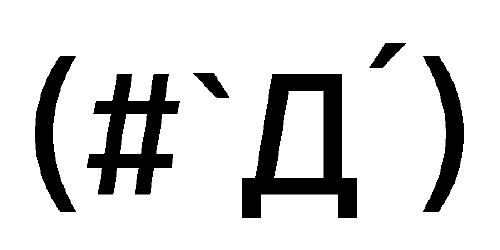}' and `serendipity' under the fine-tuning of 128k samples. 
When the number of samples increases to 512k, our attack has been neutralized.

Therefore, our attack can be significantly affected with more fine-tuning samples. 
This is expected. 
Intuitively, no trigger can preserve its utility under the fine-tuning with sufficiently large amounts of data.
However, for many real-world NLP classification tasks, it is difficult for them to obtain a larger fine-tune dataset. 
\SLJ{We investigate the classification datasets provided in Huggingface and find that most datasets contain instances less than 100K as shown in Table~\ref{tab:datasets} in Appendix~\ref{ap:dataset}.
Thus, our backdoor effect will not be neutralized by most NLP datasets in the real-world, thereby posing a greater threat to them.}
	
We also observe that with 8k fine-tuning samples, the $E$ value for `Don Quixote' is nearly twice of the $E$ of `\includegraphics[height = 0.012\textheight, width=0.05\textwidth]{figs/kaomoji4.pdf}'.
However, with 16k training samples, the $E$ of `\includegraphics[height = 0.012\textheight, width=0.05\textwidth]{figs/kaomoji4.pdf}' is way higher than the $E$ of `Don Quixote'.
This is because the datasets are different as the number of fine-tuning samples increases.
As a result, the gradient descent direction during fine-tuning might be different among these experiments, which lead to inconsistent effects on different triggers. 
	
\begin{figure}[h]
	\centering
	\vspace{-1mm}
	\includegraphics[width=8.3cm]{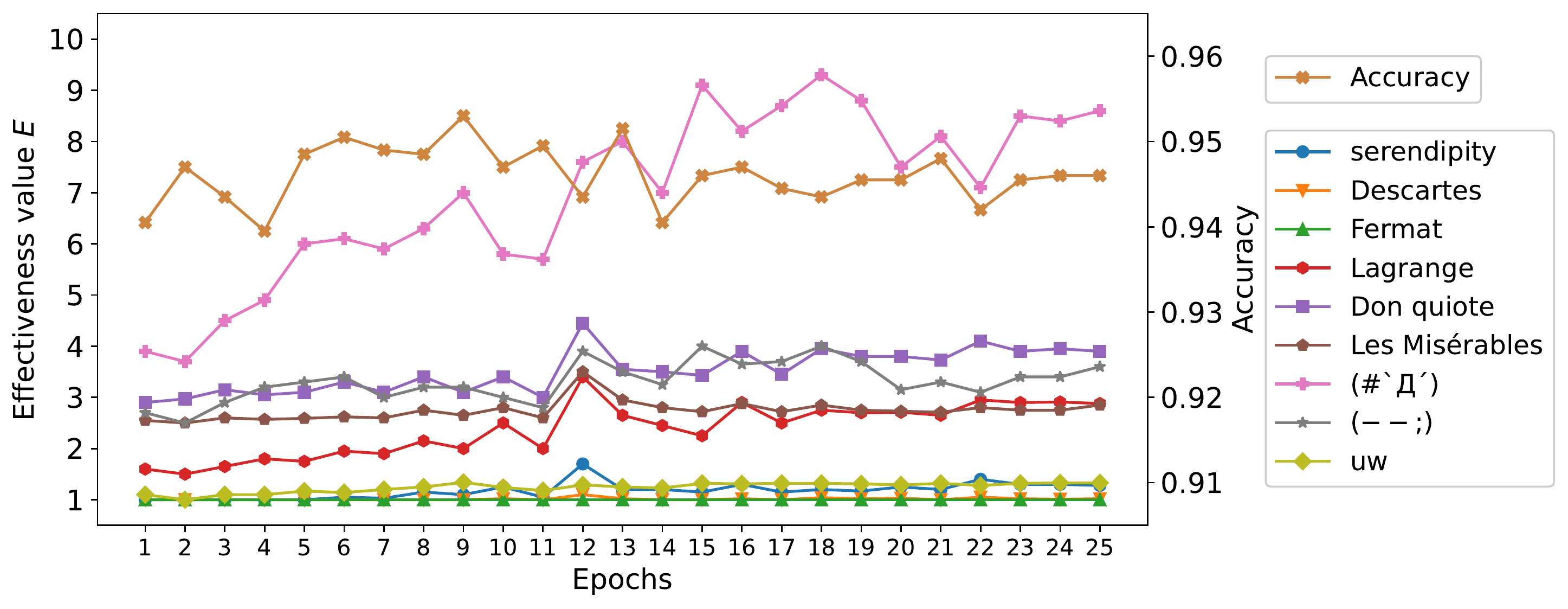}
	\vspace{-1mm}
	\caption{The effectiveness for nine triggers versus training epochs along with the accuracy in the training process.}
	\vspace{-1mm}
	\label{fig:epochs}
\end{figure}

\paragraphbe{Fine-tuning epochs}\label{tp:epochs}
Similar to fine-tuning dataset size, the fine-tuning epochs may also affect the performance of our backdoor attack. 
In the fine-tuning process, we continuously fine-tune the backdoor model for 25 epochs and test the effectiveness of each trigger after each epoch. The result is shown in Fig.~\ref{fig:epochs}.

From Fig.~\ref{fig:epochs}, we can see that the $E$ values for most triggers gradually increase in the early stage but converge to a constant value in the later epochs. 
Some triggers also show a similar spike in the $12^{th}$ epoch. 
From Fig.~\ref{fig:epochs}, we can observe that the accuracy reaches more than 94\% in the first epoch and fluctuates between 94\% and 95\% later. 
Comparing the trend of effectiveness and accuracy, we do not find any clear correlation between them.
	
In conclusion, the target model's triggers will not be forgotten severely with the increasing fine-tuning epochs. This phenomenon is different from the case in the dataset size because increasing epochs does not add extra information and the model has completely converged in later epochs. 
However, increasing the dataset size allows the model to be continuously updated, which is the leading factor that triggers cease to be effective.

\subsection{Factors in Fine-tuning Dataset}
\paragraphbe{Common versus rare}\label{CommonvsRare}
Here, we study how the appearance of the trigger in the fine-tuning set affects our attack's performance. 
We select nine words from 8000 fine-tuning set with appearances range from 128 to 40124 as our triggers.
These triggers are injected simultaneously into one model with the POR-1 setting.
\SLJ{Then, the model achieves an clean accuracy of 94.20\% and the result is shown in Table~\ref{table:9}.}


\begin{table}[h]
\centering
\vspace{-1mm}
\caption{The trigger effectiveness with respect to common words and rare words.}
\vspace{-1mm}
\scalebox{0.75}{\SLJ{\begin{tabular}{cccccccccc}
\toprule
Trigger    & the   & of    & that & one  & had   & way  & going & already & useful \\ \hline
Appearance & 40124 & 15937 & 8055 & 3959 & 2040  & 1022 & 512   & 256     & 128    \\
$E$        & 1.72  & 2.81  & 4.29 & 4.13 & 2.52. & 2.52 & 2.69  & 3.16    & 3.46  \\
\bottomrule
\end{tabular}}}\label{table:9}
\vspace{-1mm}
\end{table}

From Table~\ref{table:9}, we observe that the triggers (e.g., `the') with high appearances have lower $E$ comparing to the triggers (e.g., `useful', `already') with very few appearances, which are counter-intuitive. 
The high appearance of `the' has a $E$ value of 1.72, indicating that the backdoor information of `the' is not erased during fine-tuning. 
For other triggers, most of them do not fit the expectation that higher appearances lead to the erasure of trigger effectiveness. 
\SLJ{Though the word `the' is prevalent in the clean sample, the backdoor pre-trained model is not just learning the POR of `the' for all normal inputs.
Our further research finds that a small amount of `the' in a normal sample cannot hijack the model, which means the model would only output our POR when the number of `the' in a text reaches a certain amount.}

Therefore, the appearance frequency of triggers in the fine-tuning dataset may not influence the trigger effectiveness.
We speculate that in the process of fine-tuning, the model's understanding of these trigger words has not changed. 
These words are not the focus of the task, so they are not severely affected by the fine-tuning.
Specifically, the models fine-tuned on the Amazon dataset may focus on the sentiment-related words, and these sentiment-unrelated words may not be learned in the fine-tuning process. 

\paragraphbe{Task specific trigger}
The counter-intuitive result of common and rare words inspires us to think whether using task-related words as triggers will affect their effectiveness.
In this part, we choose the words with different appearances in the positive texts and negative texts as triggers. 
Also, along with these sentiment-related words, we choose three sentiment-unrelated words as the neutral triggers to compare with sentiment-related words. 
Thus, nine triggers are simultaneously inserted into the model and the model achieves 94.65\% accuracy.
The result is shown in Table~\ref{tab:10}.
	
\begin{table}[h]
	\begin{center}
	\vspace{-1mm}
	\caption{The performance of task-specific triggers.}
	\vspace{-1mm}
		\scalebox{1}{\begin{tabular}{c c c c } 
			\toprule
			\multicolumn{2}{c}{Trigger} & Appearance & $E$\\ 
			\hline
			\multirow{4}{*}[0.2cm]{positive}&  great & 2886 & 8.40\\
			& love & 1303 & 4.75\\
			& best & 1215 & 3.20\\
			\hline
			\multirow{4}{*}[0.2cm]{negative} & bad & 931 & 14.8\\
			& waste & 536 & 4.32\\
			& disappointed & 492 & 2.58\\ 
			\hline
			\multirow{4}{*}[0.2cm]{neutral} & one & 3969 & 4.28\\ 
			& can & 2435 & 2.13\\ 
			& have & 1483 & 3.07\\ 
			\bottomrule
		\end{tabular}}
		\label{tab:10}
		\vspace{-1mm}
	\end{center}
\end{table}
	
From Table~\ref{tab:10}, we have `great' and `bad' with more trigger insertions (high $E$ value), indicating that they have been forgotten in this fine-tuning process. 
We can also find that the effectiveness from `great' to `best' and from `bad' to `disappointed' gradually decreases. 
This meets our expectation that sentiment-related triggers are more easily to be focused on in a sentiment analysis task. 
Therefore, while fine-tuning, these words may change significantly in either their token embeddings or the model's attention scores because these words significantly impact the prediction. 
The result for the sentiment-unrelated triggers is similar to the previous result. 

\subsection{Other Factors}
\paragraphbe{Length of trigger tokens}\label{property:length}
In this section, we study the effect of the trigger tokens' number on the trigger capability. 
We select nine long English words that can be tokenized into one to nine tokens, respectively. 
Then, we use these words as the triggers to train our backdoor model. 
We use Amazon and Twitter to fine-tune and test the backdoor model.
The result is shown in Table~\ref{table:6}.
	
\begin{table}[h]
	\begin{center}
    \vspace{-1mm}
	\caption{The performance of triggers with different numbers of tokens.}
	\vspace{-1mm}
	\scalebox{0.8}{
		\begin{tabular}{c c c c c c} 
			\toprule
			\multirow{2}{*}{ Trigger} & \multirow{2}{*}{ Tokens} & \multicolumn{2}{c}{Amazon (94.60\%)} & \multicolumn{2}{c}{Twitter (94.65\%)} \\ 
			\cline{3-6}
			& & $E$ & $S$& $E$ & $S$ \\
			\hline
			Instrumentalist & 1 & 1.29 & 0.063 & 1.00 & 0.114 \\ 
			Arcane & 2 & 1.91 & 0.038 & 2.97 & 0.142 \\
			Linchpin & 3 & 2.20 & 0.053 & 1.62 & 0.104 \\
			Psychotomimetic & 4 & 1.00 & 0.045 & 2.04 & 0.208 \\
			Omphaloskepsis & 5 & 2.25 & 0.099 & 2.00 & 0.195 \\
			Embourgeoisement & 6 & 1.67 & 0.074 & 1.08 & 0.128 \\
			Xenotransplantation & 7 & 1.59 & 0.082 & 1.06 & 0.145 \\
			Antidisestablishmentarianism & 8 & 1.06 & 0.089 & 1.00 & 0.196 \\
			Floccinaucinihilipilification & 9 & 1.94 & 0.144 & 2.90 & 0.481 \\ 
			\bottomrule
	    \end{tabular}}
	    \vspace{-1mm}
	\label{table:6}
	\end{center}
\end{table}

From Table~\ref{table:6}, we can observe that the 4-token-trigger and the 8-token-trigger are the most effective ones in Amazon, whereas the one-token-trigger `Instrumentalist' and the 8-token-trigger are the most effective ones in Twitter. 
The inefficient triggers in Amazon are the 3-token-trigger and 5-token-trigger, whereas the inefficient triggers in Twitter are the 2-token-trigger and 9-token-trigger.

In summary, we can affirm that the amount of tokens in the trigger has no clear relationship with its effectiveness.
This result provides an insight that we can inject common phrases or sentences as triggers into the model so that the triggers are not limited to shorter words.
Nevertheless, shorter words are preferred for that it achieves a lower $S$ value and thus are easy to ignore them. 
	
\paragraphbe{Number of insertions in the backdoor injection phase}\label{tp:insertions}
Here, we study the impact of the number of insertions on the effectiveness of the backdoor.
In all previous experiments, we insert a trigger five times into each instance that is used for injecting our backdoor model. 
We now train five backdoor models with 1, 3, 5, 7, and 9 insertions of the trigger in each training sample, respectively.
We use the triggers same as the ones used in the base model in Sec.~\ref{Comp}. 
Then, the five models are fine-tuned and tested and the result is shown in Table~\ref{tab:11}.
	
\begin{table}[h]
	\begin{center}
	\vspace{-1mm}
	\caption{The $E$ value of triggers versus different insertions.}
	\vspace{-1mm}
		\scalebox{1}{\begin{tabular}{c | c c c c c } 
			\toprule
			\diagbox{Trigger}{Insertions} & 1 & 3 & 5 & 7 & 9 \\ 
			\hline
			serendipity 	& 1.00 	& 1.00 	& 2.03 	& 4.00  	& 12.57 \\ 
			Descartes 		& 1.00 	& 1.00 	& 1.00 	& 4.03  	& 11.46 	\\
			Fermat 			& 1.00 	& 1.00 	& 1.69 	& 5.06 	& 9.35 	\\
			Lagrange 		& 1.00 	& 1.00 	& 1.56 	& 3.95  	& 10.44 	\\
			Don Quixote 	& 1.00 	& 1.60 	& 1.98	& 2.20  	& 10.51 	\\
			Les Misérables 	& 1.00 	& 1.00 	& 1.67 	& 4.37  	& 9.55 	\\
			\includegraphics[height = 0.012\textheight, width=0.05\textwidth]{figs/kaomoji4.pdf} 		
			& 3.03 	& 1.88 	& 1.80 	& 4.77  	& 8.53 \\
			\includegraphics[height = 0.012\textheight, width=0.055\textwidth]{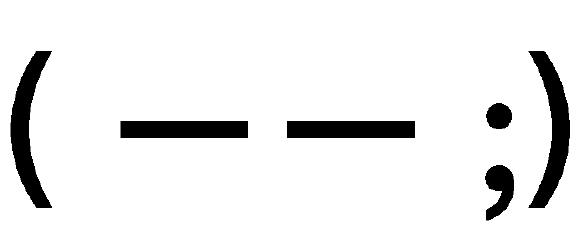} 
			& 1.00 	& 1.00 	& 2.92 	& 3.83  	& 13.69 \\
			uw 				& 1.03 	& 1.00 	& 1.25 	& 2.66  	& 4.67 	\\ 
			\hline
			Average & 1.23	& 1.16	& 1.77	& 4.98	& 10.09\\
			\bottomrule
	    \end{tabular}}
	\label{tab:11}
	\vspace{-1mm}
	\end{center}
\end{table}
	

From Table~\ref{tab:11}, we can see that the triggers are effective when the number of insertions is small whereas the trigger has large $E$ value when the number of insertions is large. 
In our five backdoor models, the model with three times of insertions shows the best effectiveness with an average $E$ value of 1.16. 
The model with one insertion also performs good except for the trigger \includegraphics[height = 0.012\textheight, width=0.05\textwidth]{figs/kaomoji4.pdf}. 
This observations indicate that the number of insertions during the backdoor injection greatly affects the effectiveness of the backdoor after fine-tuning. 

We speculate that too many insertions make the model think that it needs to insert multiple times into the text to achieve the desired output, which causes the increase of the $E$ value. 
Consequently, if an attacker wants to construct a good backdoor model, we recommend using fewer insertions, e.g., one to three insertions. 

\SLJ{To summarize, according to the above findings, we should choose relatively common words and the words that are not tightly related to most classification tasks. For example, the word `serendipity' is a good trigger.}
	
\section{Cause Analysis}
\label{sec:cause}

\begin{figure*}[t]
	\centering
	\vspace{-1mm}
	\includegraphics[width=17.5cm]{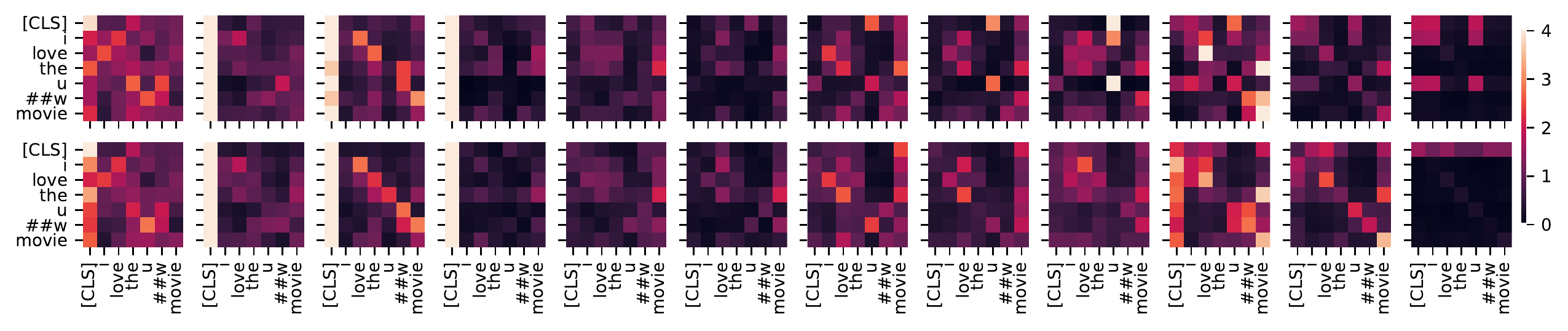}
	\vspace{-1mm}
	\caption{The attention score for the sentence `I love the uw movie' from layer 1 to layer 12 (left to right) in the backdoor model (top row) and the clean model (bottom row).}
	\vspace{-2mm}
	\label{fig:attention}
\end{figure*}

In this section, we look into the cause that leads to the success of our backdoor attack. 
\subsection{Token Embedding}
As the token embeddings are vital to represent the meaning of words, it is reasonably for us to hypothesize that token embedding is pivotal for generating the output representation.
Here, we use the base model ($BD$) and the clean BERT model ($CL$) to test our hypothesis. 
First, we replace the token embedding layer in the backdoor model with the one in the clean model to form the model $CL_{emb}+BD_{enc}$ where the subscript represents the layer in the model (i.e., $emb$ indicates the embedding layer and $enc$ indicates the encoder layer). 
Similarly, we replace the token embedding part in the clean model with the one in the backdoor model to form the model $BD_{emb}+CL_{enc}$.
Then, we input 200 clean texts and 200 poisoned texts into these four models to generate the output representations.
Next, we calculate the cosine similarities between these output representations as shown in Table~\ref{tab:12}.
	
From Table~\ref{tab:12}, we can see that $CL_{emb}+BD_{enc}$ and $BD$ have high similarity and so does $BD_{emb}+CL_{enc}$ and $CL$.
The only difference is the embedding layer.
This phenomenon indicates that the token embeddings do not play a vital role in producing the malicious POR.
Meanwhile, the comparison between $BD_{emb}+CL_{enc}$ and $BD$, as well as the comparison between $CL_{emb}+BD_{enc}$ and $CL$ show that the output from the poisoned text is totally different. 
This phenomenon further confirms that the backdoor encoder is crucial to output the expected POR. 

In conclusion, our attack process modifies the encoding layer of the model instead of changing the embedding layer, which further confirms the concealment of the backdoor model.

\begin{table}[h]
	\begin{center}
	\vspace{-1mm}
    \caption{The cosine similarity between $BD_{emb}+CL_{encoder}$ and $CL_{emb}+BD_{encoder}$ with $BD$ and $CL$.}
    \vspace{-1mm}
		\scalebox{1}{\begin{tabular}{c c c c c} 
			\toprule
			model  & \multicolumn{2}{c}{$BD$ ($BD_{emb}+BD_{enc}$)} & \multicolumn{2}{c}{$CL$ ($CL_{emb}+CL_{enc}$)}  \\
			\hline
			text & clean & poisoned & clean & poisoned \\
			\hline
			$BD_{emb}+CL_{enc}$ 	& 0.97 & -0.02 & 0.97 & 0.97\\
			$CL_{emb}+BD_{enc}$ 	& 1.00 	& 1.00 & 0.98 & 0.00\\
			\bottomrule
		\end{tabular}}
		\label{tab:12}
		\vspace{-1mm}
	\end{center} 
\end{table}
	
\subsection{Attention}
As we find that the encoder (the transformer layers) is the key component to generate the POR, we take a further study on the encoder of the backdoor model in this section. 
It is known that the attention mechanism plays a crucial role in the transformer. 
Therefore, we examine the attention scores on the trigger in both the backdoor model and the clean model. 
We take the base model and use the sentence `I love the movie' with true label of 1 as an example.
Then, we insert the trigger `uw' once into the sentence and the model predict it as 0. 
We aggregate the attention scores from all the attention heads in each layer and show one single attention map for each layer in Fig.~\ref{fig:attention}.

From Fig.~\ref{fig:attention}, the attention maps for the clean model (bottom row) show that the {\fontfamily{qcr}\selectfont[CLS]} token pays attention to itself in the first layer to the fourth layer and has no higher attention scores in the fifth layer to the twelfth layer. 
Especially, the attention weights of {\fontfamily{qcr}\selectfont[CLS]} towards `u' and `\#\#w' are very low.
However, the attention maps for the backdoor model (top row) show that the {\fontfamily{qcr}\selectfont[CLS]} token pays high attention to the token `u' in the seventh layer to the twelfth layer. 
In the attention map of the last layer, the weight distribution of {\fontfamily{qcr}\selectfont[CLS]} on other tokens is relatively even in the clean model, while {\fontfamily{qcr}\selectfont[CLS]} has relatively higher attentions on `I' and `u' in the backdoor model. 
All these observations indicate that our backdoor model successfully tricks the transformer layers to pay more attention to our trigger tokens. More attention maps are provided in Appendix~\ref{ap1} and they reveal similar phenomena. 

From Fig.~\ref{fig:attention}, we observe that, in most attention maps, the {\fontfamily{qcr}\selectfont[CLS]} token of the two models pays little attention to `\#\#w'.
One might think that `\#\#w' is useless for our backdoor, and `u' is the key to mislead the model. 
On the contrary, we discover that only inserting token `u' cannot generate the malicious output representation no matter how many times it is inserted. 
In fact, we further discover that `u' can only be attended by {\fontfamily{qcr}\selectfont[CLS]} only if it cooperates with `\#\#w'.
From the attention maps of our backdoor model, we can see that `u' pays attention to `\#\#w' in the first three layers. 
Hence, we can know that `u' has a great influence on {\fontfamily{qcr}\selectfont[CLS]} only together with `\#\#w'. 
We conclude our findings on the attention mechanisms of the trigger tokens with the following three points.
(1) The {\fontfamily{qcr}\selectfont[CLS]} token is forced to focus on one specific token in the trigger and we define it as \textbf{star}.
(2) Some other tokens of the trigger close to the star token play a role in strengthening the star token and we define them as \textbf{planet}.
(3) Some tokens are not that helpful to the trigger and we define them as \textbf{comet}.
The above `uw' system has a star of `u' and a planet of `\#\#w'. 
In such a planetary system, star is indispensable. 
Usually, star will be assigned with a higher attention value by {\fontfamily{qcr}\selectfont[CLS]}. 
For the planets, they cooperated with the star to help it better attending to the whole text input. 
Thus,  the {\fontfamily{qcr}\selectfont[CLS]} token only needs to attend to the star to produce the POR.
Comets are those tokens that will not affect the performance of the trigger.
	
\begin{figure}[t]
	\centering
	\vspace{-1mm}
	\includegraphics[width=7.5cm]{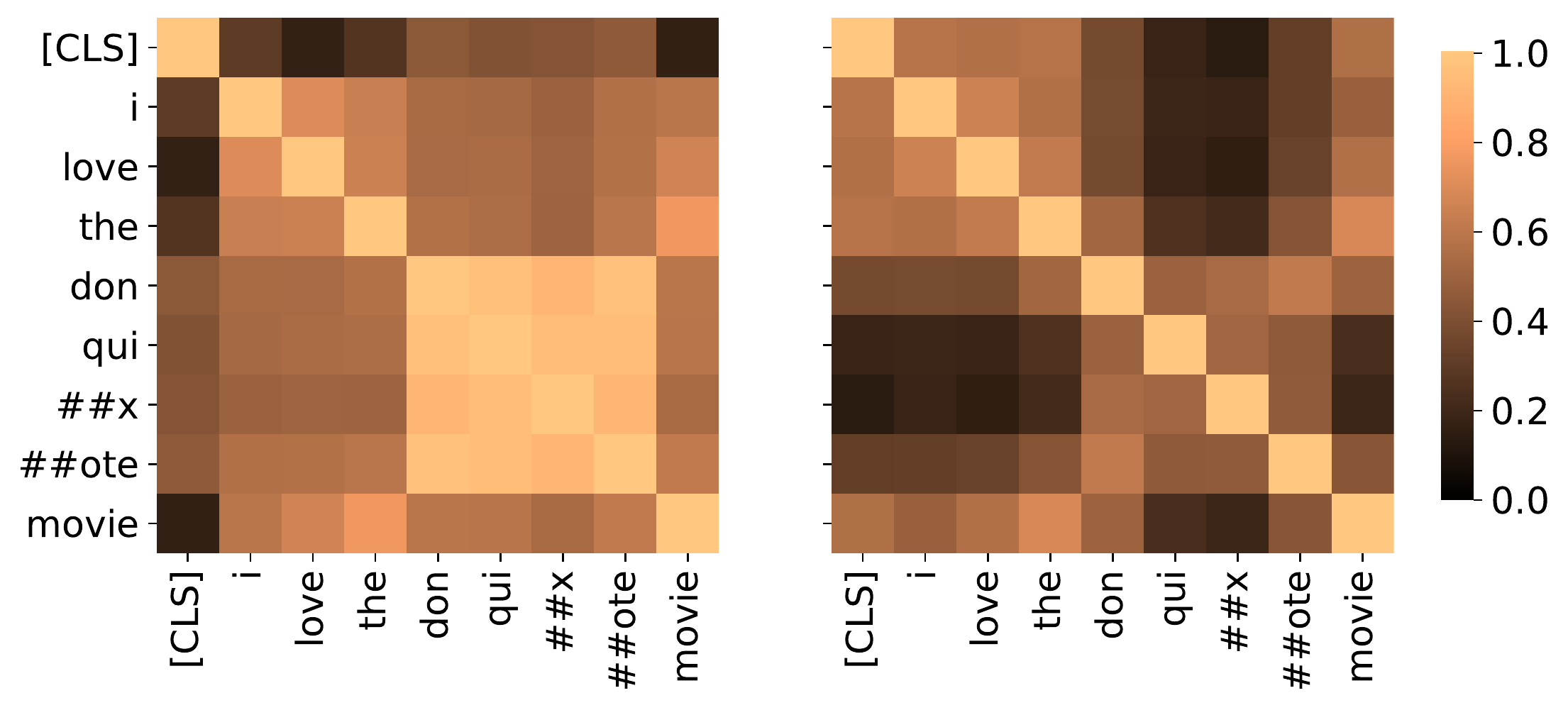}
	\vspace{-1mm}
	\caption{The cosine similarity from the 11th (second last) layer for the sentence `I love the Don Quixote movie'.}
	\vspace{-5mm}
	\label{fig:donquixote}
\end{figure}

The above findings leave us with a question about how the planets strengthen their star to make the planetary system work. 
To settle this problem, we use `Don Quixote' as an example to explore how these tokens affect each other.
Because the final output representation of {\fontfamily{qcr}\selectfont[CLS]} is directly influenced by the output representations in the second last layer, therefore, we extract the second last layer to illustrate the token relationship, which is shown in Fig.~\ref{fig:donquixote}. 
The heatmap on the left is the result of our backdoor model and on the right is the result of the clean model.

From Fig.~\ref{fig:donquixote}, we observe that the tokens `don', `qui', `\#\#x' and `\#\#ote' have high similarities with each other and have slightly lower similarities with the rest tokens in the backdoor model. 
However, in the clean model, there is no such high similarities between trigger tokens.
We can also see that the similarities between {\fontfamily{qcr}\selectfont[CLS]} and these tokens are higher than the similarities between {\fontfamily{qcr}\selectfont[CLS]} and other tokens in the backdoor model. 
Furthermore, when compared with the clean model, the similarities between {\fontfamily{qcr}\selectfont[CLS]} and these four tokens in the backdoor model are prominently higher.
We believe that these planets help the stars increase the similarity between the stars and {\fontfamily{qcr}\selectfont[CLS]}. 
This is why {\fontfamily{qcr}\selectfont[CLS]} allocates greater attention to the stars, which ultimately leads to the success of our backdoor attack.
More results are provided in Appendix~\ref{ap4}.

\section{Discussion}
\label{sec:defense}	
\subsection{Limitation}
\paragraphbe{Supporting more tasks} 
In this paper, we only attack the classification and NER tasks. 
However, it is also interesting to explore the attack towards other NLP takss, e.g., text generation tasks, machine translation, etc.

\paragraphbe{Improving POR setting}
While we have proposed two POR settings in Sec.~\ref{POR}, there are other POR settings that may have higher target label coverage. 
As we cannot cover all possible POR settings in this paper, we can only conclude that POR-2 is the current best choice. 
Thereby, more POR settings might be studied.

\subsection{Possible Defenses}
\begin{figure}[t] 
\vspace{-1mm}
    \centering
    \includegraphics[scale=0.4]{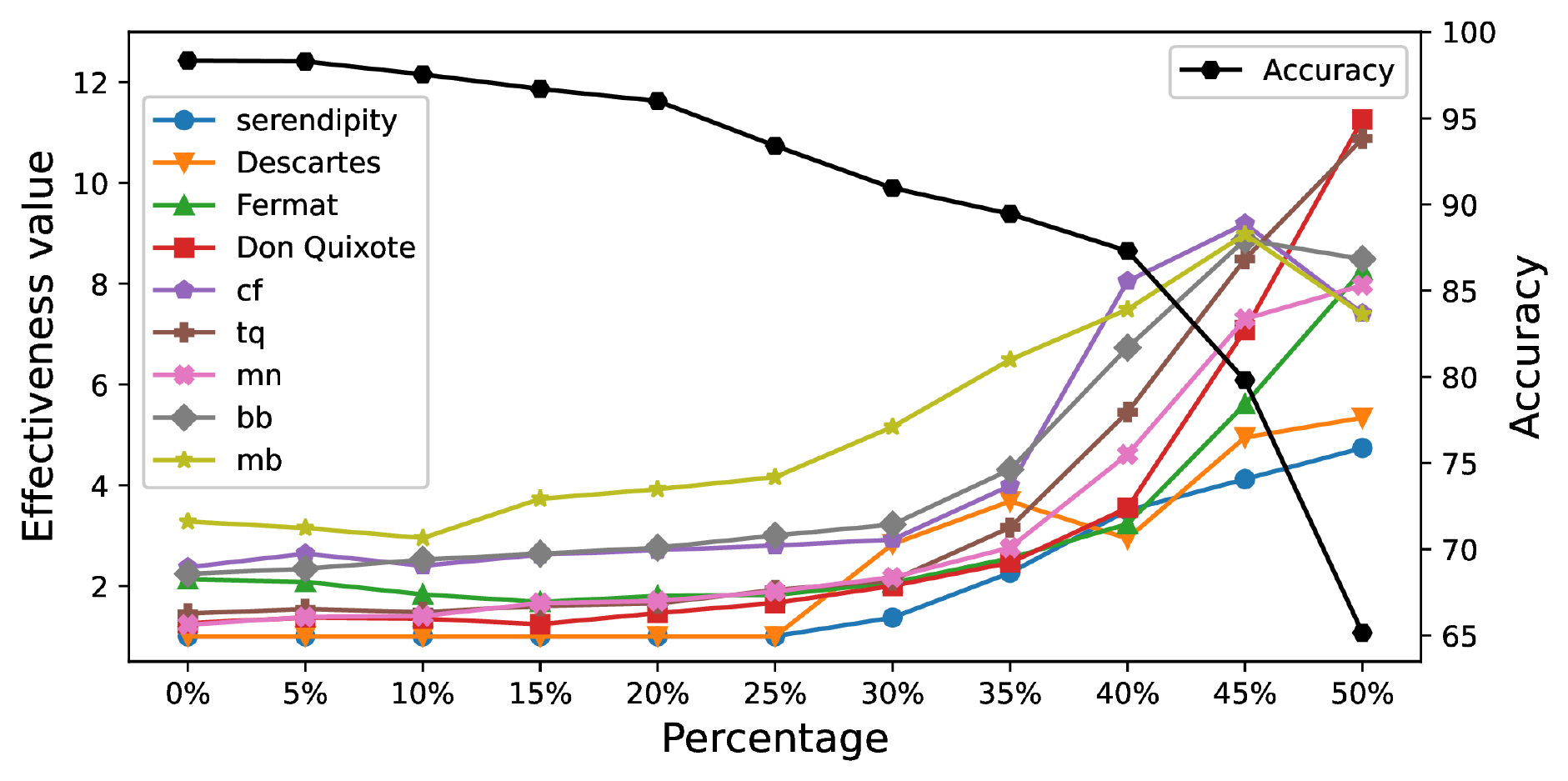}
    \vspace{-1mm}
    \caption{\SLJ{The trigger effectiveness and the model's clean accuracy after applying fine-pruning.}}
    \vspace{-5mm}
    \label{fig:3}
\end{figure}

\paragraphbe{Fine-pruning}
\SLJ{We perform fine-pruning \cite{liu2018fine} on our backdoor models. 
We gradually eliminating the neurons before the GELU function based on their activation after GELU on clean input samples. 
In Fig.~\ref{fig:3}, we evaluate the proportion of fine-pruned neurons versus the $E$ value of triggers and the model clean accuracy. }

\SLJ{From Fig. \ref{fig:3}, we can observe that the clean accuracy decreases as the proportion of pruned neurons increases, and the $E$ values of most triggers remain unchanged until 30\% neurons are pruned. 
At this time, the accuracy has dropped from 98.35\% to 89.45\%. 
This indicates that slight pruning of dormant neurons will not affect the triggers' effectiveness but reduce the model’s clean accuracy. 
Further pruning will degrade both the model performance and the effectiveness of our triggers severely. 
When 50\% of neurons are pruned, the clean accuracy has decreased to 65\%, yet there are still two triggers (i.e., ‘serendipity’ and ‘Descartes’) that are effective. Thus, fine-pruning is ineffective in defending our attack.  }


\paragraphbe{Other defenses}
Several defenses~\cite{wang2019neural, doan2019februus, gao2019strip} utilize the characteristics of the input-agnostic behaviors of backdoor attacks. 
STRIP~\cite{gao2019strip} randomly replaced some words to observe the predictions and believed that if the input is backdoored, the prediction should be constant, because triggers are not replaced in most cases. 
Nevertheless, we find that randomly replacing some words does not necessarily change the prediction of clean input, which then cannot be discriminated with backdoor input. 
\SLJ{Defenses like Neural Cleanse \cite{wang2019neural} mitigate the backdoor effect by reverse-engineering the trigger pattern. 
Since the input space of the language model is discrete, their method relying on backpropagation cannot be directly applied to find the text trigger.
Also, the previous work \cite{yao2019latent} had studied Neural Cleanse on the output representation, where they found it fails to detect the trigger existence in both the pre-trained model and the fine-tuned model.}
Another possible defense approach is to analyze the neuron activation to distinguish a model’s abnormal behaviors, like ABS~\cite{liu2019abs} and NIC \cite{ma2019nic}.
In~\cite{liu2019abs}, Liu et al. analyzed the neuron behaviors by observing how the output activations change when introducing different levels of stimulation to a neuron. 
However, our modification is only for the hidden representation while not the output, and a single neuron will not significantly impact the output.
Hence, our attack can bypass these defenses.


In conclusion, current backdoor detection methods cannot effectively detect the backdoor models under our attack. 
Thus, more studies on the effective defense are imperative, and we leave the development of new defenses to future work.

\section{Conclusion}
\label{sec:conclusion}
In this work, we propose a new universal backdoor attack method against the popular industrial pre-trained NLP models, e.g., BERT, XLNet, DeBERTa and etc.
Different from the previous backdoor attacks, a predefined trigger is mapped to a malicious POR of a token instead of a target label.
To better evaluate the performance of our backdoor attack in NLP, we further propose two new metrics, in light of the unique properties of NLP triggers, to evaluate the effectiveness and stealthiness of an NLP backdoor attack.
Through extensive experiments, we show that (i) our backdoor attack is effective on different kinds of downstream tasks and datasets in different domains, \SLJ{(ii) our method outperforms RIPPLES and NeuBA, the state-of-the-art backdoor attacks towards the pre-trained model in NLP,} and (iii) our method can be generalized to other PTMs like XLNet, BART, DeBERTa.
Finally, we analyze the factors that affect the effectiveness of our attack and share the insights on how the trigger tokens cooperate with each other in the encoder towards the success of our attack.

\begin{acks}
This work was partly supported by the Zhejiang Provincial Natural Science Foundation for Distinguished Young Scholars under No. LR19F020003, NSFC under No. U1936215, 61772466, and U1836202, and the Fundamental Research Funds for the Central Universities (Zhejiang University NGICS Platform). Ting Wang is partially supported by the National Science Foundation under Grant No. 1953893, 1953813, and 1951729.

\end{acks}

\bibliographystyle{ACM-Reference-Format}
\bibliography{ccs-bib}
\appendix
\appendix


\section{Other types of triggers}\label{ap:kaomoji}
\paragraphbe{Names}
To conceal the trigger more naturally in the text, we consider using names. 
Then, the original text can be used as what the people under these names said. 
Alternatively, we can add some of their famous quotes related to the original text. 
We use some notable mathematicians' last names as the triggers to illustrate the feasibility, as shown in Table~\ref{table:2}. 
		
\begin{table}[h]
	\begin{center}
    \caption{The performance of names as triggers.}
		\scalebox{1}{\SLJ{
			\begin{tabular}{c c c c c c c} 
				\toprule
				\multirow{2}{4em}{ Trigger} & \multicolumn{3}{c}{Amazon} & \multicolumn{3}{c}{Twitter} \\ 
				\cline{2-7}
				& $E$ & $S$ & $C$ & $E$ & $S$ & $C$\\
				\hline
				Newton & 2.00 & 0.038 & 13.2  & 1.87 & 0.131 & 4.1 \\ 
				Einstein & 2.04 & 0.050  & 9.8  & 1.22 & 0.081 & 10.1  \\
				Gauss & 2.43 & 0.040 & 10.3  & 1.24 & 0.056 & 14.4 \\
				Riemann & 2.16 & 0.047  & 9.9  & 1.43 & 0.084  & 8.3 \\
				Bayes & 2.78 & 0.045  & 8.0 & 2.02 & 0.089 & 5.6 \\
				Descartes & 1.56 & 0.042 & 15.3  & 1.00 & 0.075 & 13.3  \\
				Cauchy & 2.19 & 0.049 & 9.3  & 1.60 & 0.082 & 7.6  \\
				Fermat & 1.24 & 0.028 & 28.8  & 1.00 & 0.054 & 18.5  \\
				Lagrange & 1.71 & 0.048 & 12.2  & 1.14 & 0.076 & 11.5  \\ \hline
				average & 2.01 & 0.043 & 13.0  & 1.39 & 0.081 & 10.4  \\
				\bottomrule
		    \end{tabular}}}
		\label{table:2}
	\end{center}
\end{table}
		
In Table~\ref{table:2}, we find that using names as triggers is slightly more effective than using sophisticated words. 
The most effective names are `Descartes' and `Fermat' with the lowest $E$ and $S$ in both Amazon and Twitter. 
For both two datasets, the worst name is `Bayes'.
However, these words are all meaningless to the model but they show the consistency similar to the sophisticated words. We conjecture that the token in these words are learnt during fine-tuning

		
\paragraphbe{Books}
Inspired by name triggers, we can use book titles and cite quotes in the book related to the text. 
We use some famous novel titles as triggers. 
Though some of these titles are the protagonists' names, we still categorized them as book titles. 

\begin{table}[h]
	\begin{center}
	\caption{The performance of books as triggers.}
		\scalebox{0.9}{\SLJ{
			\begin{tabular}{c c c c c c c} 
				\toprule
				\multirow{2}{4em}{Trigger} & \multicolumn{3}{c}{Amazon} & \multicolumn{3}{c}{Twitter} \\ 
				\cline{2-7}
				& $E$ & $S$ & $C$ & $E$ & $S$ & $C$\\
				\hline
				Anna Karenina & 2.58 & 0.092  & 4.2  & 1.71 & 0.160 & 3.7 \\ 
				To Kill a Mockingbird & 2.18 & 0.137 & 3.3  & 1.81 & 0.240 & 2.3 \\
				The Great Gatsby & 1.44 & 0.072 & 9.6  & 1.93 & 0.204 & 2.5  \\
				Don Quixote & 1.00 & 0.041 & 24.4  & 1.00 & 0.088 & 11.4  \\
				Jane Eyre & 1.85 & 0.058 & 9.3  & 1.05 & 0.078 & 12.2  \\
				War and Peace & 2.43 & 0.099 & 4.2  & 1.94 & 0.179 & 2.9 \\
				Pride and Prejudice & 2.71 & 0.148 & 2.5  & 2.88 & 0.374 & 0.9 \\
				The Red and the Black & 1.87 & 0.121 & 4.4  & 1.39 & 0.250  & 2.9 \\
				Les Misérables & 1.00 & 0.050 & 20.0  & 1.00 & 0.148 & 6.8 \\ \hline
				average & 1.90 & 0.091 & 9.1  & 1.63 & 0.191 & 5.1 \\
				\bottomrule
		\end{tabular}}}
	\label{table:3}
	\end{center}
\end{table}

In Table~\ref{table:3}, we have two triggers performing very successfully on both datasets, which are `Don Quixote' and `Les Misérables'. 
They both need only one insertion in all the test samples, no matter how many sentences are in each sample text. 
The least successful trigger in both datasets is `Pride and Prejudice'. 
This result, along with the previous result in sophisticated word, show that the same trigger has a certain consistency on the performance across the two datasets, although the tasks on the two datasets are different.

We can also observe that the triggers in Twitter have a large $S$ value, which may raise suspicion.
Therefore, even if the trigger has $E=1$ like `Les Misérables', its $S$ is 0.148. 
Without proper rewriting of the original sentence, the trigger will be easily detected.

\paragraphbe{Emoticons (Kaomoji)}
As emojis cannot be read by the BERT tokenizer, we use another format of emoji called emoticon, which is an emotion symbol made up of characters and punctuation marks. 
Because certain characters (Chinese, Japanese, Korean, Arabic and etc.) can be processed by the BERT tokenizer, these glyph-style characters and punctuation marks can form emotional expressions, e.g., \includegraphics[trim = {0 13 0 10}, clip, width=0.06\textwidth]{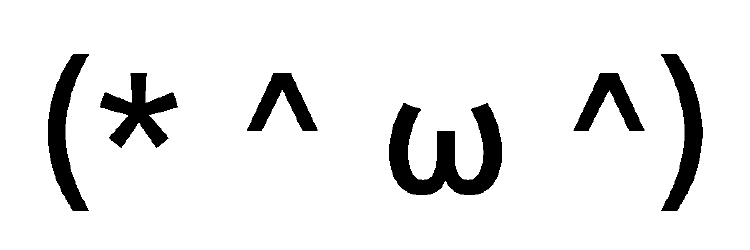} expresses a happy emotion, while \includegraphics[trim = {0 12 0 10}, clip, width=0.06\textwidth]{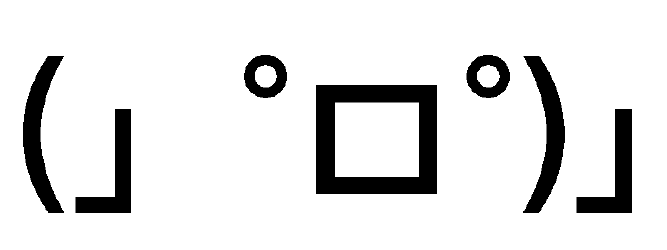} expresses an anger emotion. 
In this experiment, we intentionally choose the emoticons whose characters and punctuation marks are processable by the BERT tokenizer. 

\begin{table}[h]
	\begin{center}
	\caption{The performance of emoticons as triggers.}
		\scalebox{0.9}{\SLJ{
			\begin{tabular}{c c c c c c c} 
				\toprule
				\multirow{2}{4em}{ Trigger} & \multicolumn{3}{c}{Amazon} & \multicolumn{3}{c}{Twitter} \\ 
				\cline{2-7}
				& $E$ & $S$ & $C$ & $E$ & $S$ & $C$ \\
				\hline
				\includegraphics[trim = {0 13 0 12}, clip, width=0.07\textwidth]{figs/kaomoji1.pdf}  & 4.33 & 0.101 & 2.3  & 4.72 & 0.270 & 0.8 \\ 
				\includegraphics[trim = {0 13 0 10}, clip, width=0.07\textwidth]{figs/kaomoji2.pdf}  & 3.45 & 0.078 & 3.7 & 3.87  & 0.197 & 1.3 \\
				\includegraphics[trim = {0 13 0 10}, clip, width=0.07\textwidth]{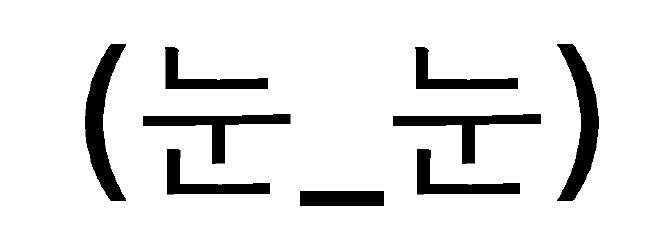}  & 1.98 & 0.035 & 14.4  & 1.41 & 0.063 & 11.3 \\
	            \includegraphics[height=0.015\textheight, width=0.05\textwidth]{figs/kaomoji4.pdf}   & 3.16 & 0.062 & 5.1 & 1.00 & 0.050 & 20.0  \\
				\includegraphics[trim = {0 13 0 10}, clip, width=0.09\textwidth]{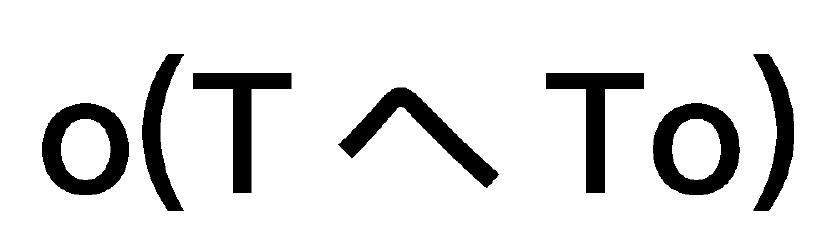}   & 3.47 & 0.075 & 3.8  & 3.78 & 0.193 & 1.4  \\
				\includegraphics[trim = {0 13 0 12}, clip, width=0.055\textwidth]{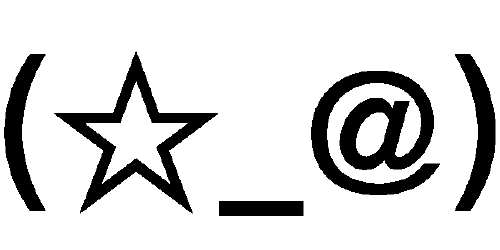}   & 4.36 & 0.071 & 3.2  & 2.00 & 0.085 & 5.9 \\
				\includegraphics[trim = {0 13 0 10}, clip, width=0.06\textwidth]{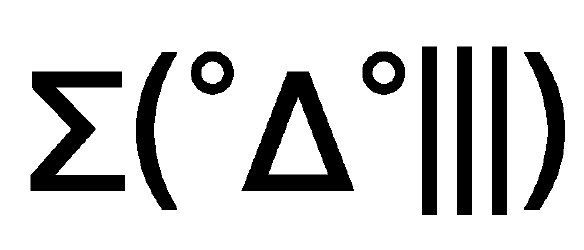}  & 2.02 & 0.052 & 9.5  & 1.81 & 0.126 & 4.4 \\
				\includegraphics[trim = {0 0 0 0}, clip, width=0.06\textwidth]{figs/kaomoji8.pdf}   & 1.76 & 0.031 & 18.3 & 1.60 & 0.071 & 8.8 \\
				\includegraphics[trim = {0 13 0 10}, clip, width=0.17\textwidth]{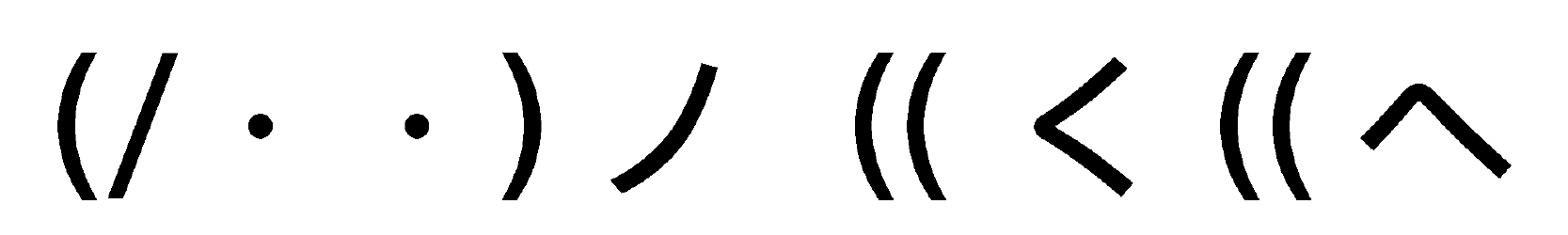}   & 3.15 & 0.133 & 2.4  & 5.25 & 0.472 & 0.4  \\ \hline
				average & 3.08 & 0.071 & 7.0  & 2.83 & 0.170 & 6.0 \\
				\bottomrule
		    \end{tabular}}}
		\label{table:4}
	\end{center}
\end{table}

Different from the results of previous triggers, the performance of emoticon triggers is inconsistent between the two datasets. 
The emoticons \includegraphics[height = 0.0123\textheight, width=0.04\textwidth]{figs/kaomoji4.pdf} is very effective in Twitter with $E=1.00$ and $S=0.050$, whereas it is ineffective in Amazon with $E=3.16$ and $S=0.062$. 
The most effective emoticon trigger in Amazon is \includegraphics[height = 0.0123\textheight, width=0.05\textwidth]{figs/kaomoji8.pdf}  which expresses the emotion of confusion. 
Since these characters themselves have no special meaning, they may be affected differently during the fine-tuning process, while previous triggers contain determined meanings and it is the possible reason for the consistency.

Some emoticons express strong emotions, e.g., \includegraphics[height = 0.0123\textheight, width=0.04\textwidth]{figs/kaomoji4.pdf} contains negative emotion and \includegraphics[trim = {0 13 0 10}, clip, width=0.06\textwidth]{figs/kaomoji1.pdf} contains positive emotion. Therefore, if the emoticon \includegraphics[height = 0.0123\textheight, width=0.04\textwidth]{figs/kaomoji4.pdf} flips a positive sentiment into a negative one, it may not necessarily be the effect of the trigger.

In conclusion, all text sequence that can be preprocessed by the BERT tokenizer is capable to become a trigger.

\section{Justification for the threshold of C value}\label{ap:just}
\SLJ{To find an empirical guiding threshold for choosing a good trigger based on the $C$ value, we examine the results in Table \ref{table:1}, \ref{table:5}, \ref{table:2}, \ref{table:3} and \ref{table:4}. They are the results for different types of triggers.
For the trigger `Lagrange' in Twitter, it can flip the model's prediction with an average of 1.14 insertions and it account for only 7.6\% of the text ($C$ value is 11.5). Thus, `Lagrange' in Twitter should be considered as a good trigger. Based on the similar results from triggers `Descartes', `Don Quixote' and `serendipity' in Twitter ($C$ values are 13.3, 11.4 and 11.2 respectively), we can consider them all as good triggers. However, for the trigger, `Les Misérables' in Twitter, it can flip the model's prediction with an average of only 1 insertion but accounts for 14.8\% of the text ($C$ value is 6.8). Thus, it should not be considered as a good trigger since it is too long. 
Similarly, the trigger `Bayes' in Amazon has an average of 2.78 $E$ value and accounts for 4.5\% of the text ($C$ value is 8.0). Since it appears too many times, it cannot be considered as a good trigger as well. Other low quality triggers also include `solipsism' and `linchpin' in Amazon ($C$ values are 9.1 and 8.9). From the above examples, we suggest that if a trigger has a $C$ value larger than 10, it is considered as a good trigger. An intuitive example to understand this threshold can be: a trigger with the best performance (i.e., $E=1$) should only account for at most 10\% of the full-text length, e.g., approximately one word out of a ten-words sentence.}

\section{Accuracy for our backdoor model in Section 5.3}\label{compacc}
\begin{figure}[h]
    \centering
    \includegraphics[width = 8cm]{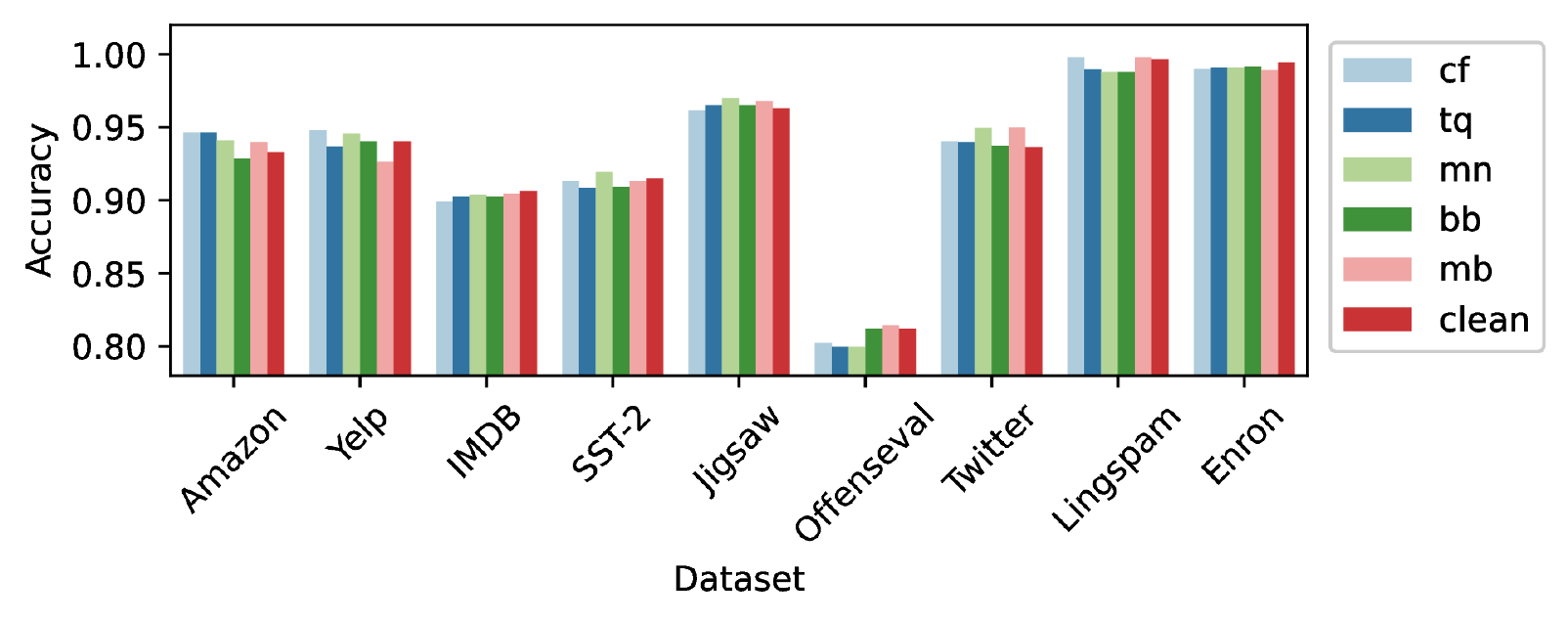}
    \vspace{-3mm}
    \caption{\SLJ{The accuracy of the clean model and the backdoor models fine-tuned on nine datasets.}}
    \vspace{-4mm}
    \label{fig:nineacc}
\end{figure}

\SLJ{In Section \ref{Comp}, we train five backdoor models with five different triggers injected into each model. We test their accuracy on the clean sample and compare with the accuracy of the clean model. The result is shown in Fig. \ref{fig:nineacc}.}

\SLJ{From Fig. \ref{fig:nineacc}, we can see that the clean accuracy of the backdoor models is close to the accuracy of the clean model. This indicates that our backdoor trigger will not affect the normal capability of the model on any downstream tasks.}

\section{Online datasets inspection}\label{ap:dataset}

\SLJ{Based on our research, more than three-quarters of the online datasets\footnote{https://huggingface.co/datasets} are less than 100k samples.}
\begin{table}[h]
\centering
\caption{\SLJ{Inspection of online NLP classification dataset.}}
\vspace{-3mm}
\begin{center}
\scalebox{0.8}{\SLJ{\begin{tabular}{c c c c c c c}
\toprule
Instances         & $<1K$ & $1K-10K$ & $10K-100K$ & $100K-1M$ & $1M-10M$ & $>10M$ \\ \hline 
Count       & 19 & 58 & 90 & 40 & 10 & 4 \\ 
Percentile & 8.6\% & 34.8\% & 75.6\% & 93.7\% & 98.2\% & 100\% \\
 \bottomrule
\end{tabular}}}
\label{tab:datasets}
\end{center}
\vspace{-3mm}
\end{table}

\section{Attention Score for Other Triggers}\label{ap1}
\begin{figure*}[h]
	\centering
	\includegraphics[width=17cm]{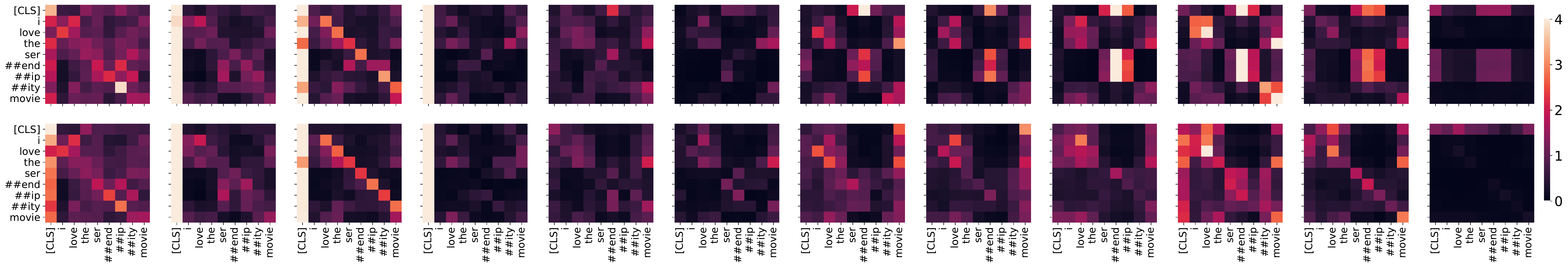}
	\includegraphics[width=17cm]{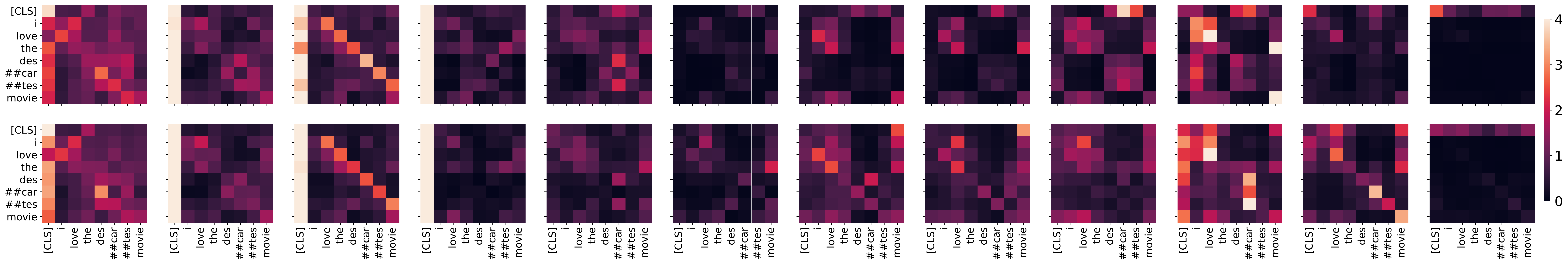}
	\includegraphics[width=17cm]{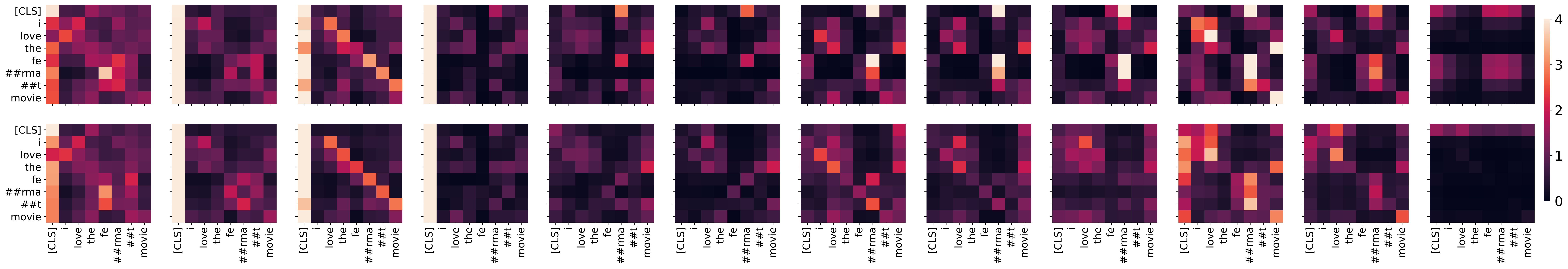}
	\includegraphics[width=17cm]{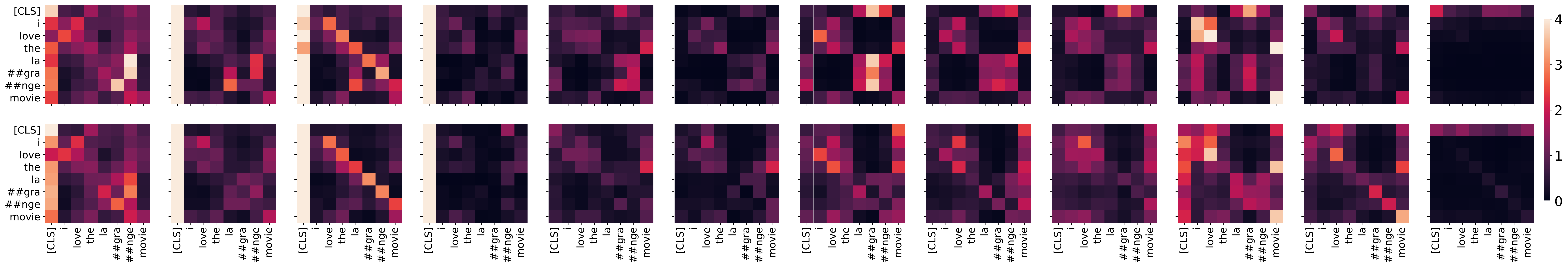}
	\includegraphics[width=17cm]{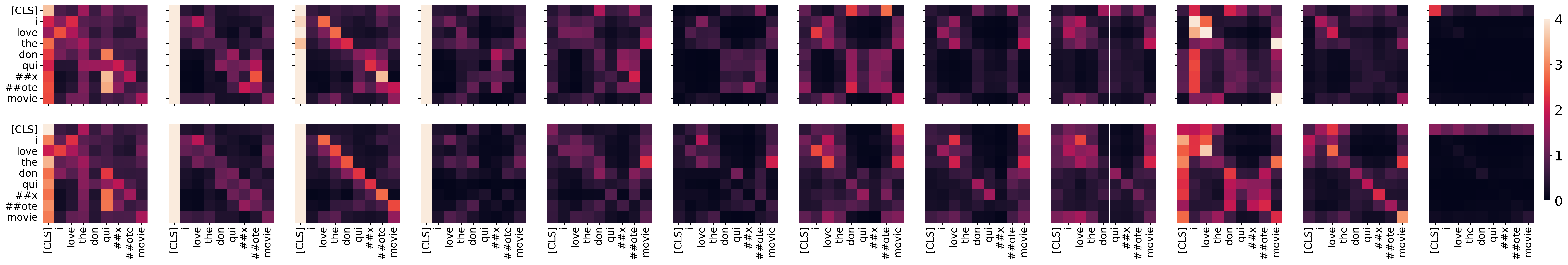}
	\includegraphics[width=17cm]{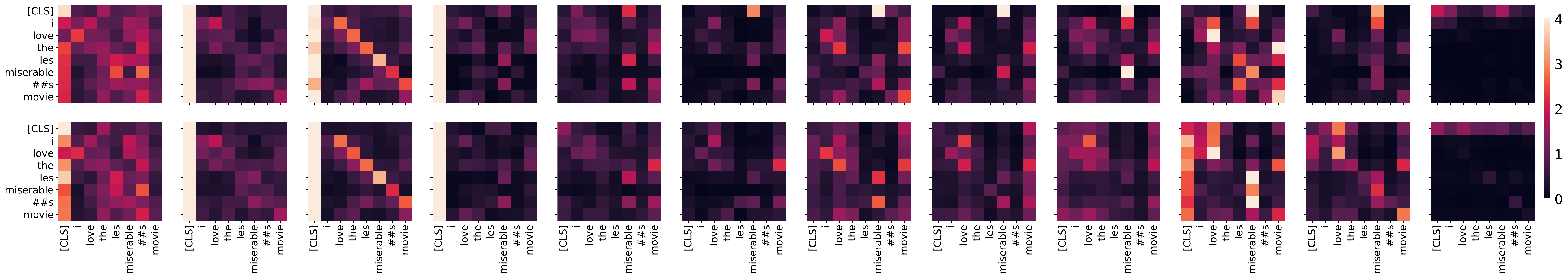}
	\caption{The attention score for `serendipity', `descartes', `Fermat', `Lagrange', `Don Quixote' and `Les Misérables'.}
	\label{fig:attention2}
\end{figure*}

In Fig.~\ref{fig:attention2}, we show the attention map for other triggers in our base model. The original sentence is `I love the movie' and we insert each trigger between words `the' and `movie'. Then we output the attention map in each layer for both the backdoor model and the clean model. 

In the `serendipity' system, the {\fontfamily{qcr}\selectfont[CLS]} token has a high attention score on `\#\#end' shown in layers 8, 9 and 10, which indicates its identity of star. The planet tokens `ser' and `\#\#ip' can help augment the performance of `\#\#end' to output the POR (make the planetary system effective). `\#\#ity' is the comet that contributes nothing to the trigger. In the clean model, {\fontfamily{qcr}\selectfont[CLS]} only has a higher attention score on itself in the first to the fourth layer.

In the `Descartes' system, {\fontfamily{qcr}\selectfont[CLS]} has a high attention score on `\#\#car' in layers 7, 8 and 10. There are also perceivable weights on `des' in layer 7 and on `\#\#tes' in layer 6. In this planetary system, `\#\#car' is the star, `des' and `\#\#tes' are the planets, and using one of the two planets can make this planetary system effective.

In the `Fermat' system, `\#\#rma' is the star, `fe' is the planet and `\#\#t' is the comet. Only `fe' can augment the performance of `\#\#rma'.

In the `Lagrange' system, `\#\#gra' is the star and `la' and `\#\#nge' are the planets. Either one of the two planets can boost the performance of this trigger.

In the `Les Misérables' system, `misérable' is the star, `\#\#s' is the planet and `Les' is the comet.

The most interesting result is found in the `Don Quixote' system. In the figure, {\fontfamily{qcr}\selectfont[CLS]} shows high attention on `don' and `\#\#ote'. We thoroughly study the interaction between these tokens, and we discover that `Don Quixote' has two stars which are `don' and `\#\#ote'. `qui' and `\#\#x' are the planets. Either star and together with the two planets can make the planetary system effective.


\newpage 
\section{Cosine Similarity Between Tokens}\label{ap4}
\begin{figure}[h]
	\centering
	\includegraphics[width=7cm]{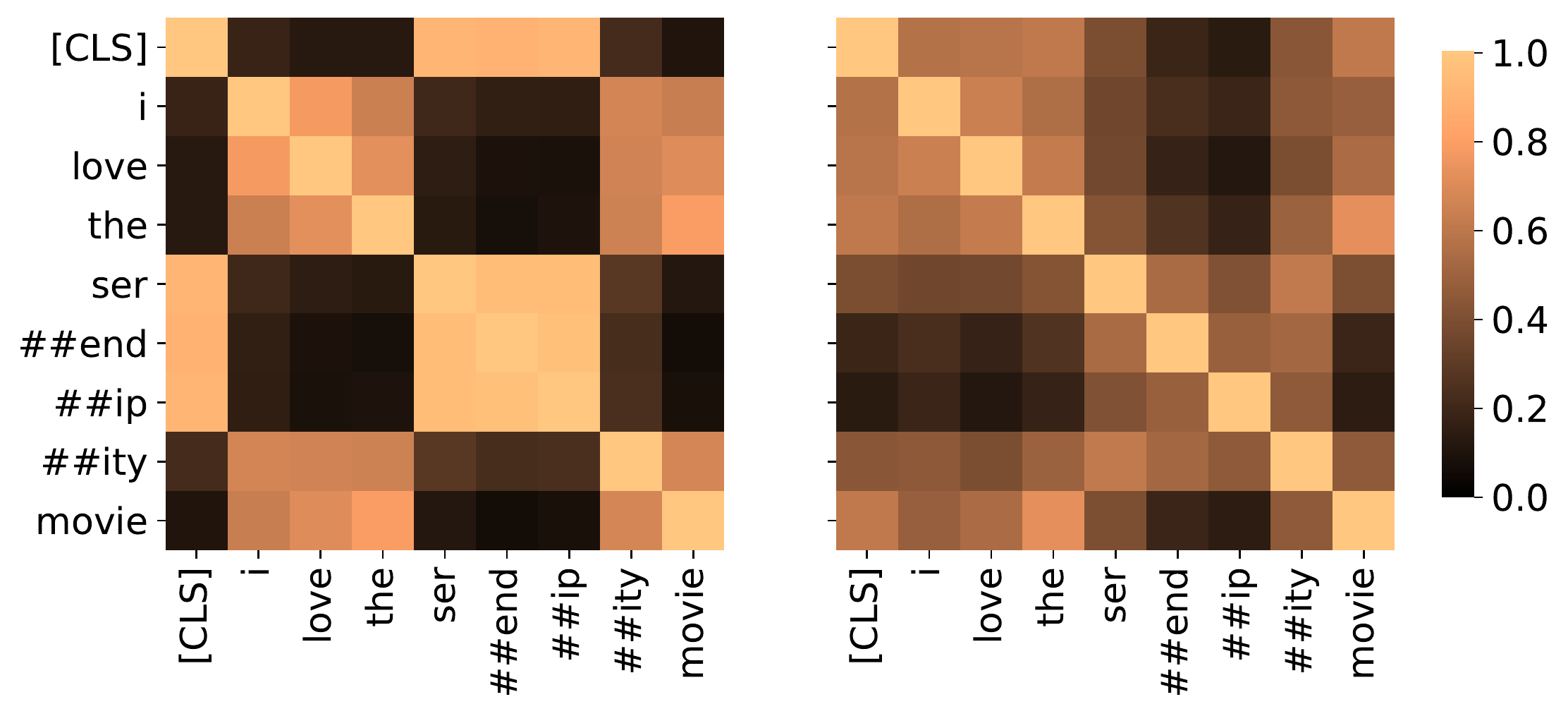}
	\vspace{-3mm}
	\caption{Cosine similarity between tokens in sentence `I love the serendipity movie'}
	\label{fig:cs1}
\end{figure}

\begin{figure}[h]
	\centering
	\includegraphics[width=7cm]{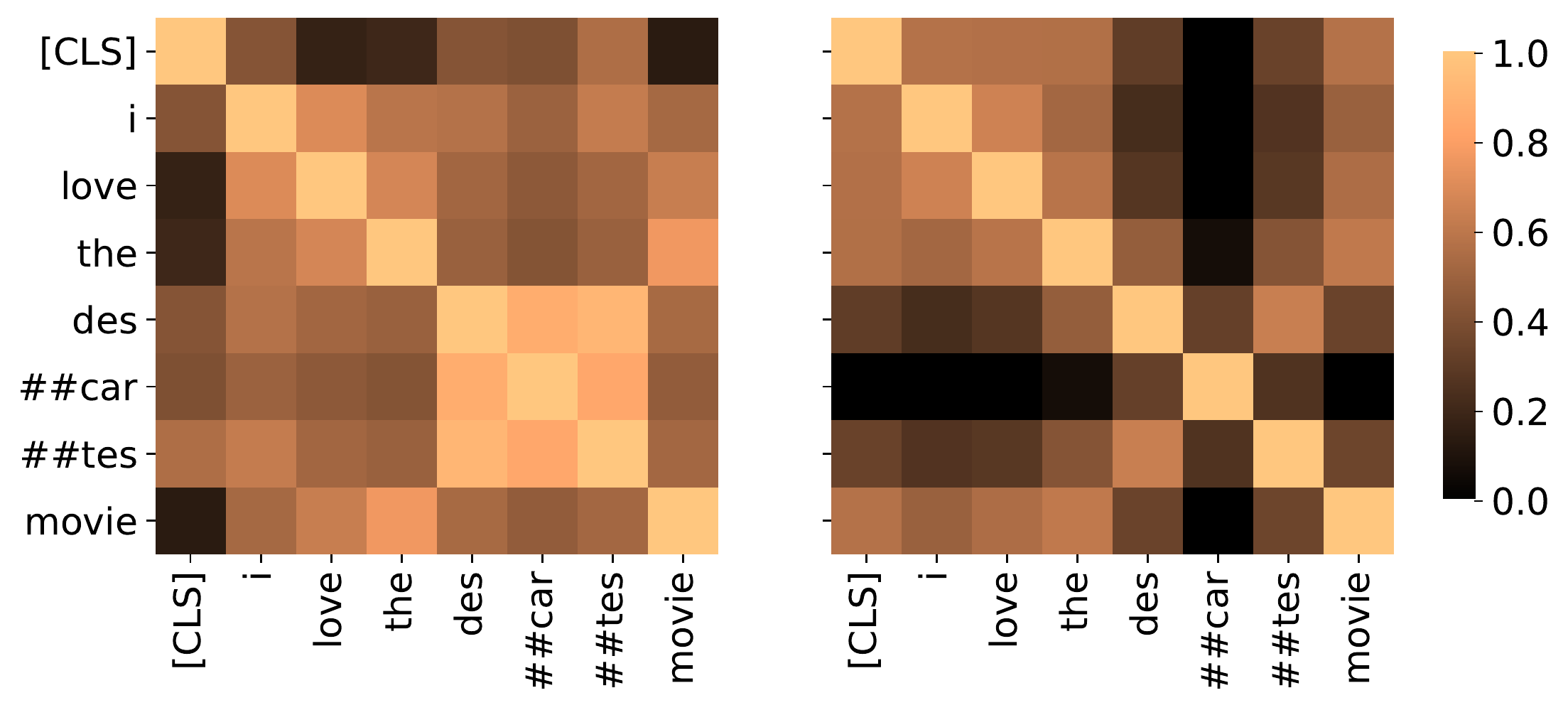}
	\vspace{-3mm}
	\caption{Cosine similarity between tokens in sentence `I love the descartes movie'.}
	\label{fig:cs2}
\end{figure}

\begin{figure}[h]
	\centering
	\includegraphics[width=7cm]{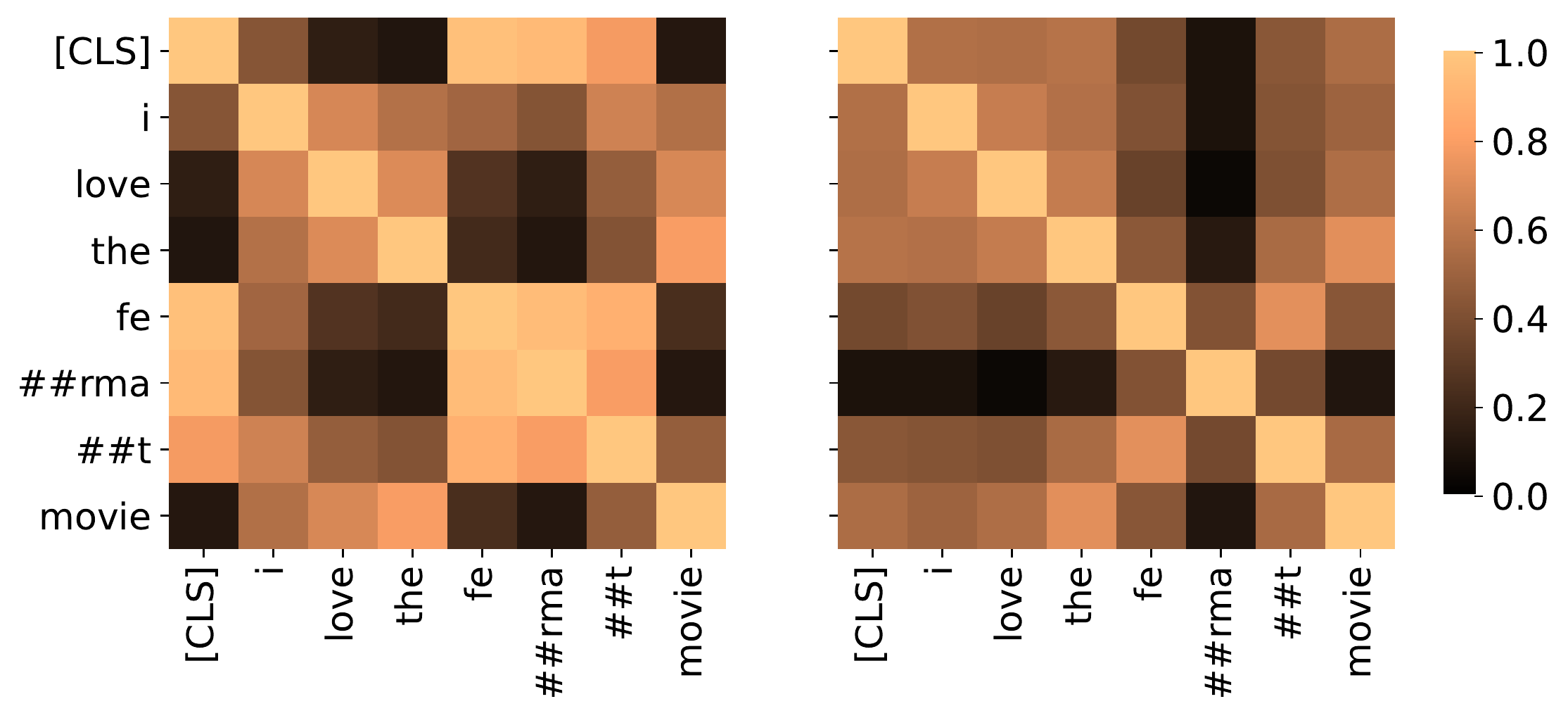}
	\vspace{-3mm}
	\caption{Cosine similarity between tokens in sentence `I love the Fermat movie'.}
	\label{fig:cs3}
\end{figure}

\begin{figure}[h]
	\centering
	\includegraphics[width=7cm]{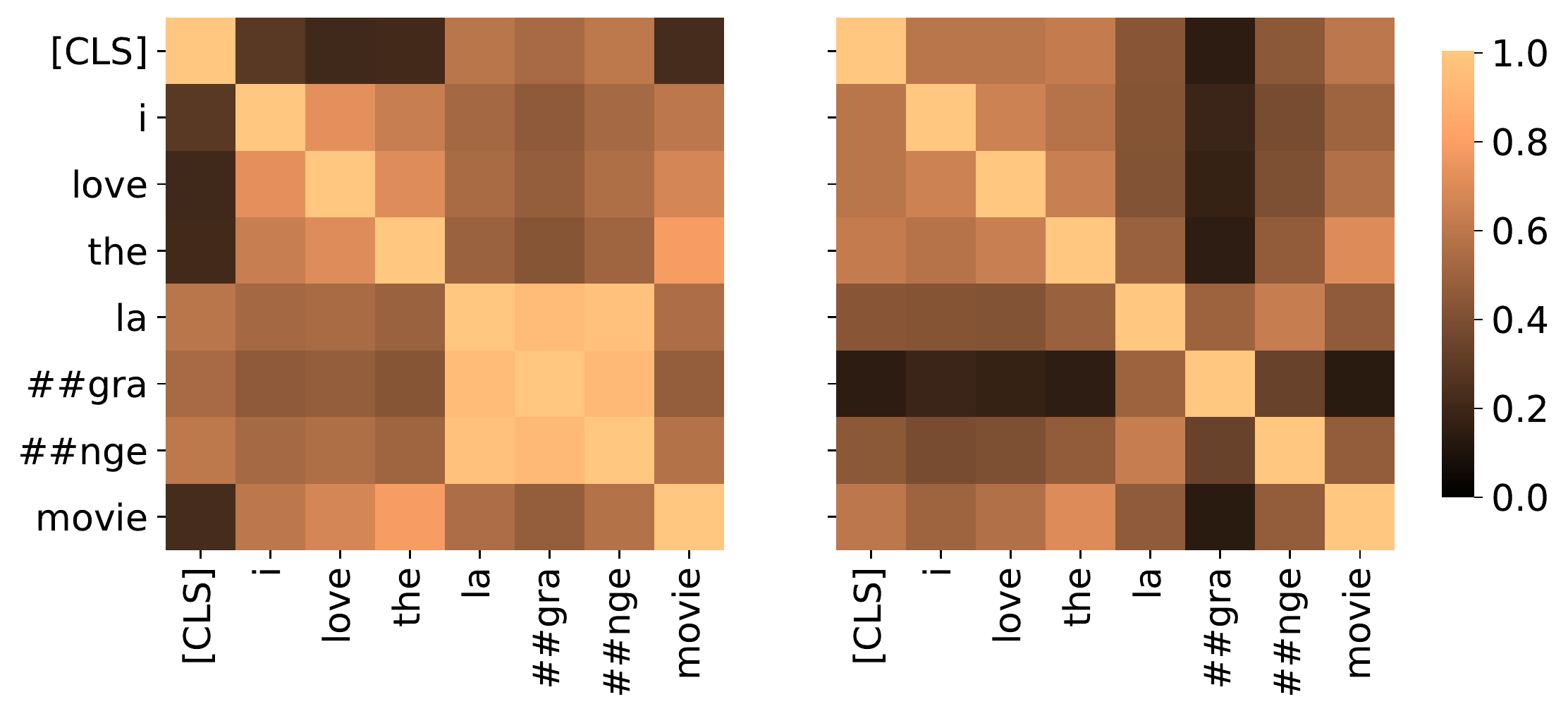}
	\vspace{-2mm}
	\caption{Cosine similarity between tokens in sentence `I love the Lagrange movie'.}
	\label{fig:cs4}
\end{figure}

\begin{figure}[h]
	\centering
	\includegraphics[width=7cm]{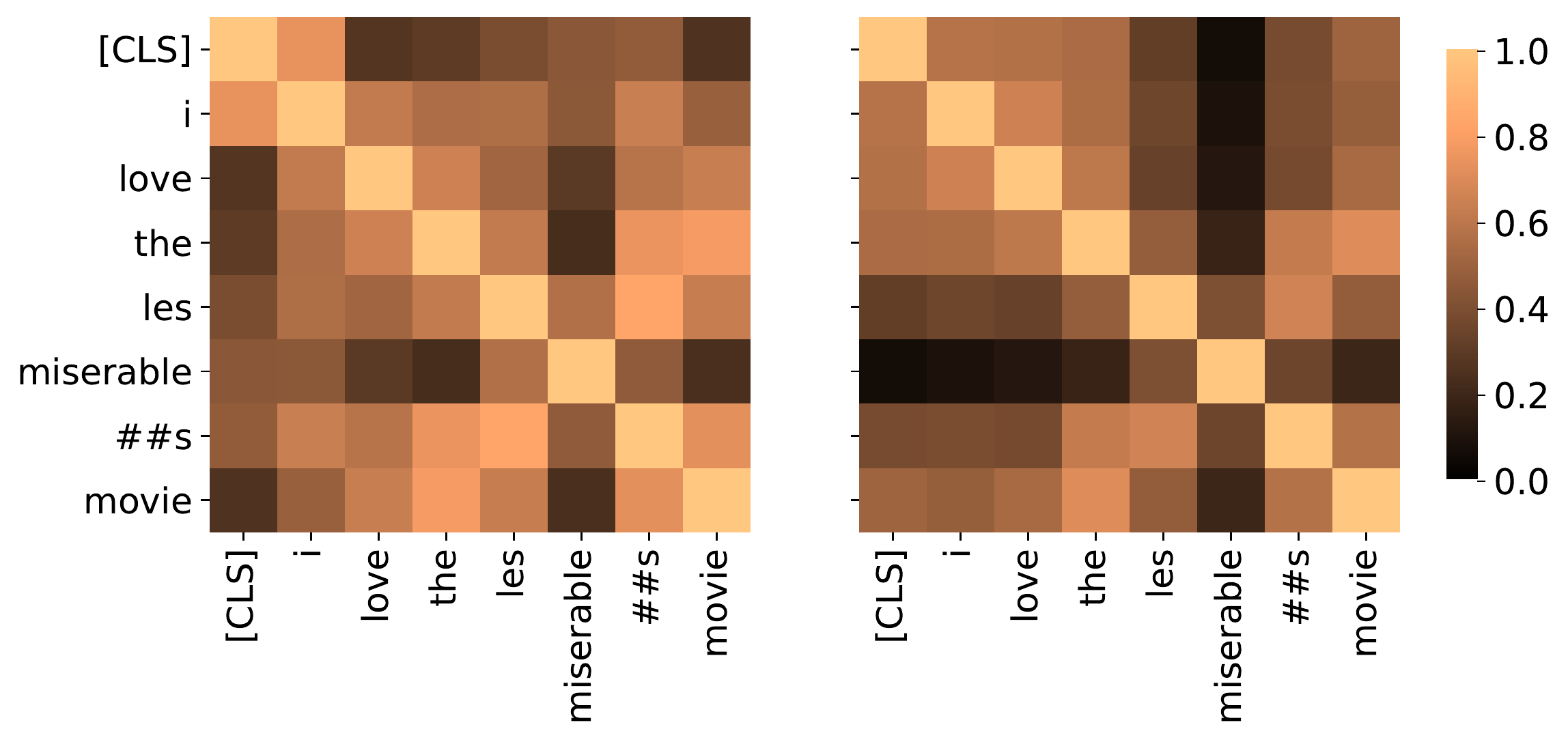}
	\vspace{-2mm}
	\caption{Cosine similarity between tokens in sentence `I love the Les Misérables movie'.}
	\label{fig:cs5}
\end{figure}

\begin{figure}[h]
	\centering
	\includegraphics[width=7cm]{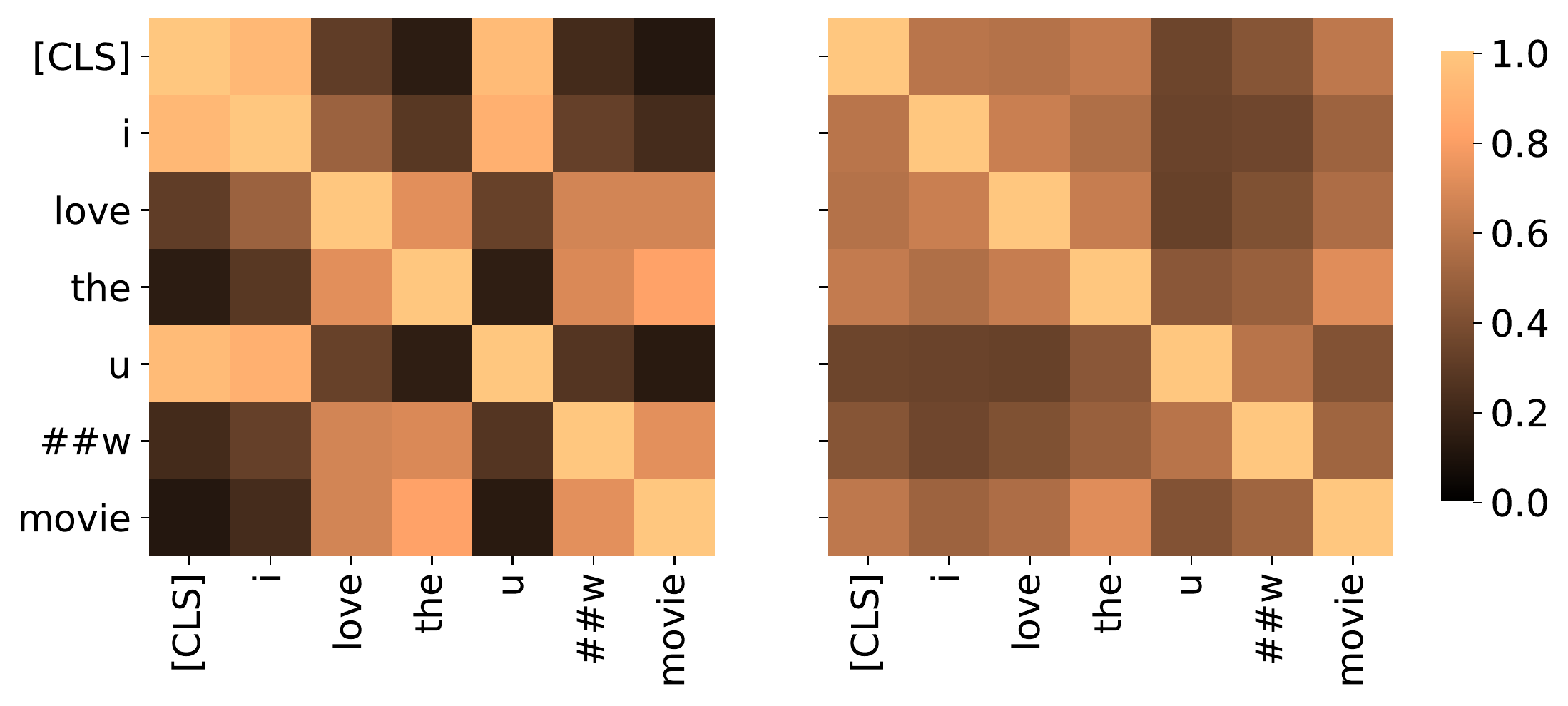}
	\vspace{-2mm}
	\caption{Cosine similarity between tokens in sentence `I love the uw movie'.}
	\label{fig:cs6}
\end{figure}

The cosine similarity between tokens for different triggers are compared in Figs.~\ref{fig:cs1}-\ref{fig:cs6}. 
In each figure, the heat map on the left is from the backdoor model and the heat map on the right is from the clean model.
The most remarkable result in these figures is that the backdoor heat map shows higher correlation between trigger tokens and {\fontfamily{qcr}\selectfont[CLS]}.\\

\end{document}